\pdfoutput=1
\documentclass[11pt]{article}

\usepackage[]{ACL2023}

\usepackage{times}
\usepackage{latexsym}

\usepackage[T1]{fontenc}
\usepackage[utf8]{inputenc}

\usepackage{microtype}

\usepackage{graphicx}
\usepackage{amsmath}
\usepackage{breqn}
\usepackage{makecell}
\usepackage{tabularx}
\usepackage{booktabs}
\usepackage{multirow}
\usepackage[colorinlistoftodos]{todonotes}
\usepackage{textcomp}
\usepackage{bm}

\usepackage{hyperref}
\usepackage{amsmath}
\usepackage{multirow}
\usepackage{pifont}
\usepackage{xspace}

\usepackage{inconsolata}
\usepackage{algorithm}
\usepackage{algpseudocode}
\usepackage{caption}
\usepackage{subcaption}
\usepackage{enumerate}
\usepackage{multicol}

\usepackage{times}
\usepackage{latexsym}

 \usepackage{amssymb}
 \usepackage{amsmath}
 \usepackage{subcaption}

\newenvironment{itemize*}%
  {\begin{itemize}%
    \setlength{\itemsep}{0pt}%
    \setlength{\parskip}{0pt}}%
  {\end{itemize}}
  \newenvironment{enumerate*}%
  {\begin{enumerate}%
    \setlength{\itemsep}{0pt}%
    \setlength{\parskip}{0pt}}%
  {\end{enumerate}}

\newcommand{\cmark}{\ding{51}}
\newcommand{\xmark}{\ding{55}}
\usepackage{makecell}

\newcommand{\benchmark}{RoSE\xspace}
\title{Revisiting the Gold Standard: \\ Grounding Summarization Evaluation with Robust Human Evaluation}

\author{
 Yixin Liu\Thanks{~~Equal contribution} $^{1}$ 
 \quad \textbf{Alexander R. Fabbri}\textsuperscript{$*$}$^{2}$
 \quad \textbf{Pengfei Liu}$^{3}$ 
 \quad \textbf{Yilun Zhao}$^{1}$ \\
 \quad \textbf{Linyong Nan}$^{1}$ 
 \quad \textbf{Ruilin Han}$^{1}$ 
 \quad \textbf{Simeng Han}$^{1}$ 
 \quad \textbf{Shafiq Joty}$^{2}$ \\
 \quad \textbf{Chien-Sheng Wu}$^{2}$ 
 \quad \textbf{Caiming Xiong}$^{2}$
 \quad \textbf{Dragomir Radev}$^{1}$ \\
  $^1$Yale University, 
  $^2$Salesforce AI,
  $^3$Carnegie Mellon University  \\
  \texttt{yixin.liu@yale.edu}, \texttt{afabbri@salesforce.com}
 }

\begin{document}
\maketitle

\begin{abstract}
Human evaluation is the foundation upon which the evaluation of both summarization systems and automatic metrics rests. 
However, existing human evaluation studies for summarization either exhibit a low inter-annotator agreement or have insufficient scale, and an in-depth analysis of human evaluation is lacking.
Therefore, we address the shortcomings of existing summarization evaluation along the following axes: 
(1) We propose a modified summarization salience protocol, \textit{Atomic Content Units} (ACUs), which is based on fine-grained semantic units and allows for a high inter-annotator agreement.
(2) We curate the Robust Summarization Evaluation (\textbf{\benchmark}) benchmark, a large human evaluation dataset consisting of 22,000 summary-level annotations over 28 top-performing systems on three datasets.
(3) We conduct a comparative study of four human evaluation protocols, underscoring potential confounding factors in evaluation setups.
(4) We evaluate 50 automatic metrics and their variants using the collected human annotations across evaluation protocols and demonstrate how our benchmark leads to more statistically stable and significant results.
The metrics we benchmarked include recent methods based on large language models (LLMs), GPTScore and G-Eval.
Furthermore, our findings have important implications for evaluating LLMs, as we show that LLMs adjusted by human feedback (e.g., GPT-3.5) may overfit unconstrained human evaluation, which is affected by the annotators' prior, input-agnostic preferences, calling for more robust, targeted evaluation methods.
\end{abstract}
\section{Introduction}\label{sec:introduction}

\newcommand{\tabdress}[1]{{{#1}}}

\begin{table*}[t!]
\small
\centering
\addtolength{\tabcolsep}{-2pt} 
\setlength{\extrarowheight}{1pt}
\begin{tabular}{c p{0.78\linewidth}}
\toprule
\textbf{Statistical Power}  &\tabdress{$-$ High statistical power is difficult to reach for human evaluation of similar-performing systems.}\\
\S\ref{subsec:power-analysis} &\tabdress{$-$ Increasing the sample size of human evaluation effectively raises statistical power.}  \\
\midrule
\textbf{Summary Length} &\tabdress{$-$ Summaries from different summarization systems show a large difference in average length.}\\
\S\ref{subsec:sys-analysis} &\tabdress{$-$ Difference in summary length is not well-reflected by automatic evaluation metrics.} \\
\midrule
 &\tabdress{$-$ Reference-free and reference-based human evaluation results have a near-zero correlation.} \\
 \textbf{Evaluation} &\tabdress{$-$ Reference-free human evaluation strongly correlates with input-agnostic, annotator preference.}\\
\textbf{Protocol Comparison}  
 &\tabdress{$-$ Annotator's input-agnostic preference has a strong positive correlation with summary lengths.} \\
 \S\ref{subsec:protocal-result-analysis} &\tabdress{$-$ Annotator's input-agnostic preference does not favor reference summaries.} \\
&\tabdress{$-$ Compared to smaller, fine-tuned models, zero-shot large language models (e.g. GPT-3) perform better under reference-free evaluation, but worse under reference-based evaluation.}\\
\midrule
 \multicolumn{1}{p{2.5cm}}{~ \par ~~~~~~~~~\textbf{Evaluating}} &\tabdress{$-$ A higher-powered human evaluation dataset can lead to a more robust automatic metric evaluation, as shown by a tighter confidence interval and higher statistical power of metric evaluation.} \\
\textbf{Automatic Metrics}  & \tabdress{$-$ Automatic metric performance differs greatly under different human evaluation protocols.} \\ 
\S\ref{subsec:metric-eval} \& \S\ref{subsec:metric-analysis} & \tabdress{$-$ Automatic metrics show relatively strong system-level correlation and moderate summary-level correlation with our robust human evaluation protocol.}\\
\bottomrule
\end{tabular}
\addtolength{\tabcolsep}{2pt} 
\setlength{\extrarowheight}{-1pt}
\vspace{-2mm}
\caption{Summary of the key findings in our work.}
\label{tab:intro} 
\vspace{-3mm}
\end{table*}

Human evaluation plays an essential role in both assessing the rapid development of summarization systems in recent years~\cite{lewis-etal-2020-bart,10.5555/3524938.3525989,NEURIPS2020_1457c0d6,sanh2022multitask,he2022z} and in assessing the ability of automatic metrics to evaluate such systems as a proxy for manual evaluation~\cite{bhandari-etal-2020-evaluating, TACL2563, gao-wan-2022-dialsummeval}.
However, while human evaluation is regarded as the gold standard for evaluating both summarization systems and automatic metrics, as suggested by \citet{clark-etal-2021-thats} an evaluation study does not become ``gold'' automatically without proper practices.
For example, achieving a high inter-annotator agreement among annotators can be difficult~\cite{goyal-gpt3}, and there can be a near-zero correlation between the annotations of crowd-workers and expert annotators~\cite{TACL2563}. Also, a human evaluation study without a large enough sample size can fail to find statistically significant results due to insufficient statistical power~\cite{card-etal-2020-little}.

\par
Therefore, we believe it is important to ensure that \textbf{human evaluation can indeed serve as a solid foundation for evaluating summarization systems and automatic metrics}. For this, we propose using a robust \textbf{\textit{human evaluation protocol}} for evaluating the salience of summaries that is more objective by dissecting the summaries into fine-grained content units and defining the annotation task based on those units.
Specifically, we introduce the \textit{Atomic Content Unit} (ACU) protocol for summary salience evaluation (\S\ref{sec:methodology}), which is modified from the Pyramid \cite{nenkova-passonneau-2004-evaluating} and LitePyramid \cite{shapira-etal-2019-crowdsourcing} protocols.
We demonstrate that with the ACU protocol, a high inter-annotator agreement can be established among crowd-workers,
which leads to more stable system evaluation results and better reproducibility.
\par
We then collect, through both in-house annotation and crowdsourcing, \textbf{\benchmark}, a large \textbf{\textit{human evaluation benchmark}} of human-annotated summaries with the ACU evaluation protocol on recent state-of-the-art summarization systems, which yields higher statistical power (\S\ref{sec:benchmark}). 
To support evaluation across datasets and domains, our benchmark consists of test sets over three summarization datasets, CNN/DailyMail (CNNDM)~\cite{Nallapati:16}, XSum~\cite{narayan-etal-2018-dont}, and SamSum~\cite{gliwa-etal-2019-samsum}, and annotations on the validation set of CNNDM to facilitate automatic metric training.
To gain further insights into the characteristics of different evaluation protocols, we conduct \textbf{\textit{human evaluation with three other protocols}} (\S\ref{sec:protocol-comparison}). 
Specifically, we analyze protocol differences in the context of both fine-tuned models and large language models (LLMs) in a zero-shot setting such as GPT-3~\cite{NEURIPS2020_1457c0d6}.
We find that different protocols can lead to drastically different results, which can be affected by annotators' prior preferences, highlighting the importance of aligning the protocol with the summary quality intended to be evaluated.
We note that our benchmark enables a more trustworthy \textbf{\textit{evaluation of automatic metrics}} (\S\ref{sec:metric-comparison}), as shown by statistical characteristics such as tighter confidence intervals and more statistically significant comparisons (\S\ref{subsec:metric-analysis}).
Our evaluation includes recent methods based on LLMs~\cite{Fu2023GPTScoreEA, Liu2023GEvalNE}, and we found that they cannot outperform traditional metrics despite their successes on related benchmarks such as SummEval~\cite{TACL2563}.

We summarize our key findings in Tab.~\ref{tab:intro}.
Our contributions are the following: 
(1) We propose the ACU protocol for high-agreement human evaluation of summary salience. 
(2) We curate the \textbf{\benchmark} benchmark, 
consisting of 22000 summary-level annotations and requiring over 150 hours of in-house annotation, across three summarization datasets, which can lay a solid foundation for training and evaluating automatic metrics.\footnote{We release our benchmark and evaluation scripts at \url{https://github.com/Yale-LILY/ROSE}.} 
(3) We compare four human evaluation protocols for summarization and show how they can lead to drastically different model preferences.
% .
(4) We evaluate automatic metrics across different human evaluation protocols and call for human evaluation to be conducted with a clear evaluation target aligned with the evaluated systems or metrics, such that \textit{task-specific} qualities can be evaluated without the impact of general, \textit{input-agnostic} preferences of annotators.
We note that the implications of our findings can become even more critical with the progress of LLMs trained with human preference feedback~\citep{ouyang2022training} and call for a more rigorous human evaluation of LLM performance.

\section{Related Work}\label{sec:related_work}

\noindent \textbf{Human Evaluation Benchmarks}
Human annotations are essential to the analysis of summarization research progress.
Thus, recent efforts have focused on aggregating model outputs and annotating them according to specific quality dimensions~\cite{huang-etal-2020-achieved,bhandari-etal-2020-evaluating,10.5555/3495724.3495977,zhang-bansal-2021-finding,TACL2563,gao-wan-2022-dialsummeval}.
The most relevant work to ours is \citet{bhandari-etal-2020-evaluating}, which annotates summaries according to semantic content units, motivated by the Pyramid \cite{nenkova-passonneau-2004-evaluating} and LitePyramid \cite{shapira-etal-2019-crowdsourcing} protocols.
However, this benchmark only covers a single dataset (CNNDM) without a focus on similarly-performing state-of-the-art systems, which may skew metric analysis \cite{Aggrefact2022} and not fully reflect realistic scenarios \cite{deutsch-etal-2022-examining}.
In contrast, our benchmark consists only of outputs from recently-introduced models over three datasets.

\noindent \textbf{Summarization Meta-Evaluation}
With a human evaluation dataset, there exist many directions of meta-evaluation, or re-evaluation of the current state of evaluation, 
such as metric performance analyses, understanding model strengths, and human evaluation protocol comparisons.
\par
Within metric meta-analysis, several studies have focused on the analysis of ROUGE \cite{lin-2004-rouge}, and its variations \cite{rankel-etal-2013-decade,graham-2015-evaluating}, across domains such as news \cite{lin2004looking}, meeting summarization~\citep{liu-liu-2008-correlation}, and scientific articles~\citep{cohan-goharian-2016-revisiting}. 
Other studies analyze  a broader set of metrics \cite{peyrard-2019-studying, bhandari-etal-2020-evaluating,deutsch2020sacrerouge,TACL2563,gabriel-etal-2021-go,kasai-etal-2022-bidimensional}, including
those specific to factual consistency evaluation \cite{kryscinski2019factual,durmus-etal-2020-feqa,wang-etal-2020-asking, maynez2020,Laban2022SummaCRN, fabbri-etal-2022-qafacteval,honovich-etal-2022-true,tam2022evaluating}.
\par 
Regarding re-evaluating model performance, a recent line of work has focused on evaluating zero-shot large language models \cite{goyal-gpt3,liang-hollistic,tam2022evaluating}, noting their high performance compared to smaller models. 

\par
As for the further understanding of human evaluation, prior work has compared approaches to human evaluation \cite{hardy-etal-2019-highres}, studied annotation protocols for quality dimensions such as linguistic quality \cite{steen-markert-2021-evaluate} and factual consistency \cite{tang-etal-2022-investigating}, and noted the effects of human annotation inconsistencies on system rankings \cite{owczarzak2012}.
The unreliability and cost of human evaluation in certain settings have been emphasized \cite{chaganty-etal-2018-price, clark-etal-2021-thats},  with some work noting that thousands of costly data points may need to be collected in order to draw statistically significant conclusions \cite{wei-jia-2021-statistical}.
Our meta-analysis focuses on this latter aspect, 
and we further analyze potential confounding factors in evaluation such as length and protocol design, with respect to both small and large zero-shot language models.
\section{Atomic Content Units for Summarization Evaluation}\label{sec:methodology}

\begin{figure*}[t!]
    \centering
         \includegraphics[width=0.98\linewidth]{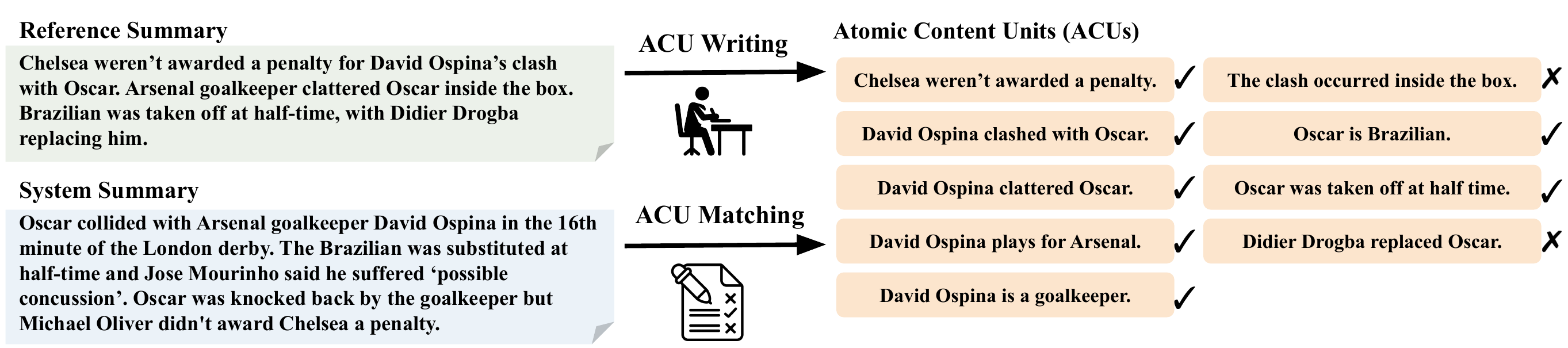}
 \caption{\label{fig:example_annotation}Example of a reference summary, a system summary and corresponding ACU annotations on CNNDM. 
    }
\end{figure*}

We now describe our Atomic Content Unit (ACU) annotation protocol for reference-based summary salience evaluation, including the procedure of writing ACUs based on reference summaries and matching the written ACUs with system outputs.
\subsection{Preliminaries}
In this work, we focus on a specific summarization meta-evaluation study on \textit{summary salience}.
Salience is a desired summary quality that requires the summary to include all and only important information of the input article.
The human evaluation of summary salience can be conducted in either \textit{reference-free} or \textit{reference-based} manners. 
The former asks the annotators to assess the summary directly based on the input article~\cite{TACL2563}, while the latter requires the annotators to assess the information overlap between the system output and reference summary~\cite{bhandari-etal-2020-evaluating}, under the assumption that the reference summary is the gold standard of summary salience.\footnote{We note salience can be an inherently subjective quality, and the reference summary of common datasets may not always be the actual ``gold standard'', discussed more in \S\ref{sec:concludsion-new}.}
Given that reference-based protocols are more constrained, we focus on \textit{\textbf{reference-based evaluation}} for our human judgment dataset collection, and we conduct a comparison of protocols in \S\ref{sec:protocol-comparison}.
\subsection{ACU Annotation Protocol}
\label{subsec:acu-protocol}
Inspired by the Pyramid~\cite{nenkova-passonneau-2004-evaluating} and LitePyramid~\cite{shapira-etal-2019-crowdsourcing} protocols and subsequent annotation collection efforts \cite{bhandari-etal-2020-evaluating,zhang-bansal-2021-finding}, the ACU protocol is designed to reduce the subjectivity of reference-based human evaluation by simplifying the basic annotation unit -- the annotators only need to decide on the presence of a single fact, extracted from one text sequence, in another text sequence, to which a binary label can be assigned with more objectivity. 
Specifically, the evaluation process is decomposed into two steps: (1) \textbf{ACU Writing} -- extracting facts from one text sequence, and (2) \textbf{ACU Matching} -- checking for the presence of the extracted facts in another sequence.
We formulate the ACU protocol as a \textit{\textbf{recall-based}} protocol, such that the first step only needs to be performed once for the reference summary, allowing for  reproducibility and reuse of these units when performing matching on new system outputs.

\noindent \textbf{ACU Writing}
While the LitePyramid approach defines its basic content unit as a sentence containing a brief fact, we follow \citet{bhandari-etal-2020-evaluating}
to emphasize a shorter, more fine-grained information unit.
Specifically, we define the ACU protocol with the concept of \textbf{\textit{atomic facts}} -- elementary information units in the reference summaries, which no longer need to be further split for the purpose of reducing ambiguity in human evaluation.\footnote{We note that it can be impossible to provide a practical definition of atomic facts. 
Instead, we use it as a general concept for fine-grained information units.}
Then, ACUs are constructed based on one atomic fact and other minimal, necessary information.

Fig.~\ref{fig:example_annotation} shows an example of the written ACUs.
To ensure annotation quality, we (the authors) write all the ACUs used in this work.
We define guidelines to standardize the annotation process; for each summary sentence the annotator creates an ACU constituting the main information from the subject of the main clause (e.g., root), followed by additional ACUs for other facts while including the minimal necessary information from the root.
We provide rules for dealing with quotations, extraneous adjectives, noisy summaries, and additional cases.
We note that there can still be inherent subjectivity in the written ACUs among different annotators even with the provided guidelines.
However, such subjectivity should be unbiased in summary comparison because all the candidate summaries are evaluated by the same set of written ACUs.

\noindent \textbf{ACU Matching}
Given ACUs written for a set of reference summaries, our protocol evaluates summarization system performance by checking the presence of the ACUs in the system-generated summaries as illustrated in Fig.~\ref{fig:example_annotation}.
For this step, we recruit annotators on Amazon Mechanical Turk\footnote{\url{https://www.mturk.com/}} (MTurk).
The annotators must pass a qualification test, and additional requirements are specified in Appendix~\ref{append:benchmark}.
Besides displaying the ACUs and the system outputs, we also provide the reference summaries to be used as context for the ACUs.

\noindent \textbf{Scoring Summaries with ACU}
ACU matching annotations can be aggregated into summary scores.
We first define an un-normalized ACU score $f$ of a candidate summary $s$ given a set of ACUs $\mathcal{A}$ as:
\begin{equation}
\label{eq:acu}
\small
    f(s, \mathcal{A}) = \frac{|\mathcal{A}_s|}{|\mathcal{A}|},
\end{equation}
where $\mathcal{A}_s$ is a subset of $\mathcal{A}$ that is matched with $s$.
We note that $f$ by default is a \textit{recall} based score with respect to the reference summary $r$.
Therefore, we also define a \textit{normalized} ACU score $\tilde{f}$ as:
\begin{equation}
\label{eq:normalzied-acu}
\small
  \tilde{f}_\alpha(s, \mathcal{A}, r) =  e^{\min{(0, \frac{1 - \frac{|s|}{|r|}}{\alpha}})}  f(s, \mathcal{A}),
\end{equation}
where $|s|$, $|r|$ are the length (i.e., number of words) of the candidate summary $s$ and the reference summary $r$ respectively, and $\alpha$ is a positive number controlling the strength of the normalization.
This normalization is in effect a \textit{redundancy penalty}, which penalizes the summaries longer than the reference and resembles the brevity penalty in BLEU~\cite{papineni-etal-2002-bleu}.
In practice, we set the value of $\alpha$ by de-correlating $\tilde{f}$ with the summary length using the collected ACU annotations.

\subsection{ACU Annotation Collection}
\label{sec:results}

\begin{table}[t!]
\small
\centering
\addtolength{\tabcolsep}{-1pt} 
\begin{tabular}{lccccc}
\toprule
\textbf{Dataset} & \textbf{Split} & \textbf{\#Doc.} & \textbf{\#Sys.}  & \textbf{\#ACU} & \textbf{\#Summ.}\\
\midrule
CNNDM & Test & 500 & 12 & 5.6k & 6k \\
CNNDM & Valid & 1,000 & 8 & 11.6k & 8k \\
XSum & Test & 500 & 8 & 2.3k & 4k \\
SamSum & Test & 500 & 8 & 2.3k & 4k\\
\bottomrule
\end{tabular}
\caption{Statistics of the collected annotations. 
\textbf{\#Doc.} is the number of input documents, \textbf{\#Sys.} is the number of summarization systems used for collection.
\textbf{\#ACU} is the total number of written ACUs.
\textbf{\#Summ.} is the total number of summary-level annotations, which are aggregated over three annotations on the test sets, and a single annotation on the validation set of CNNDM.
}
\addtolength{\tabcolsep}{1pt} 
\label{tab:data-statistics} 
\end{table}

We collect ACU annotations on three summarization datasets: CNNDM~\cite{Nallapati:16}, XSum~\cite{narayan-etal-2018-dont}, and SamSum~\cite{gliwa-etal-2019-samsum}.
To reflect the latest progress in text summarization, we collect and annotate the generated summaries of pre-trained summarization systems proposed in recent years.\footnote{We release all of the system outputs with a unified, cased, untokenized format to facilitate future research.}
Detailed information about the summarization systems we used can be found in Appendix~\ref{subsec:models-appendix}.

Table~\ref{tab:data-statistics} shows the statistics of the collected annotations.
The annotations are collected from the test set of the above datasets, and additionally from the validation set of CNNDM to facilitate the training of automatic evaluation metrics. 
In total, we collect around 21.8k ACU-level annotations and around 22k summary-level annotations, aggregated over around 50k individual summary-level judgments. 

To calculate inter-annotator agreement, we use Krippendorff's alpha~\citep{Krippendorff2011ComputingKA}.
The aggregated summary-level agreement score of ACU matching is 0.7571, and the ACU-level agreement score is 0.7528.
These agreement scores are higher than prior collections, such as RealSumm~\cite{bhandari-etal-2020-evaluating} and SummEval~\cite{TACL2563}, which have an average agreement score of crowd-workers 0.66 and 0.49, respectively.

\section{\benchmark Benchmark Analysis}
\label{sec:benchmark}
We first analyze the robustness of our collected annotations and a case study on the system outputs.
\subsection{Power Analysis}
\label{subsec:power-analysis}
We analyze the \textit{statistical power} of our collected human annotations to study whether it can yield stable and trustworthy results~\cite{card-etal-2020-little}.
Statistical power is the probability that the null hypothesis of a statistical significance test is rejected given there is a real effect.
For example, for a human evaluation study that compares the performance of two genuinely different systems, a statistical power of 0.80 means there is an 80\% chance that a significant difference will be observed.
Further details can be found in Appendix~\ref{subsec:power-details}.
\begin{figure}[t!]
    \centering
         \includegraphics[width=0.85\linewidth]{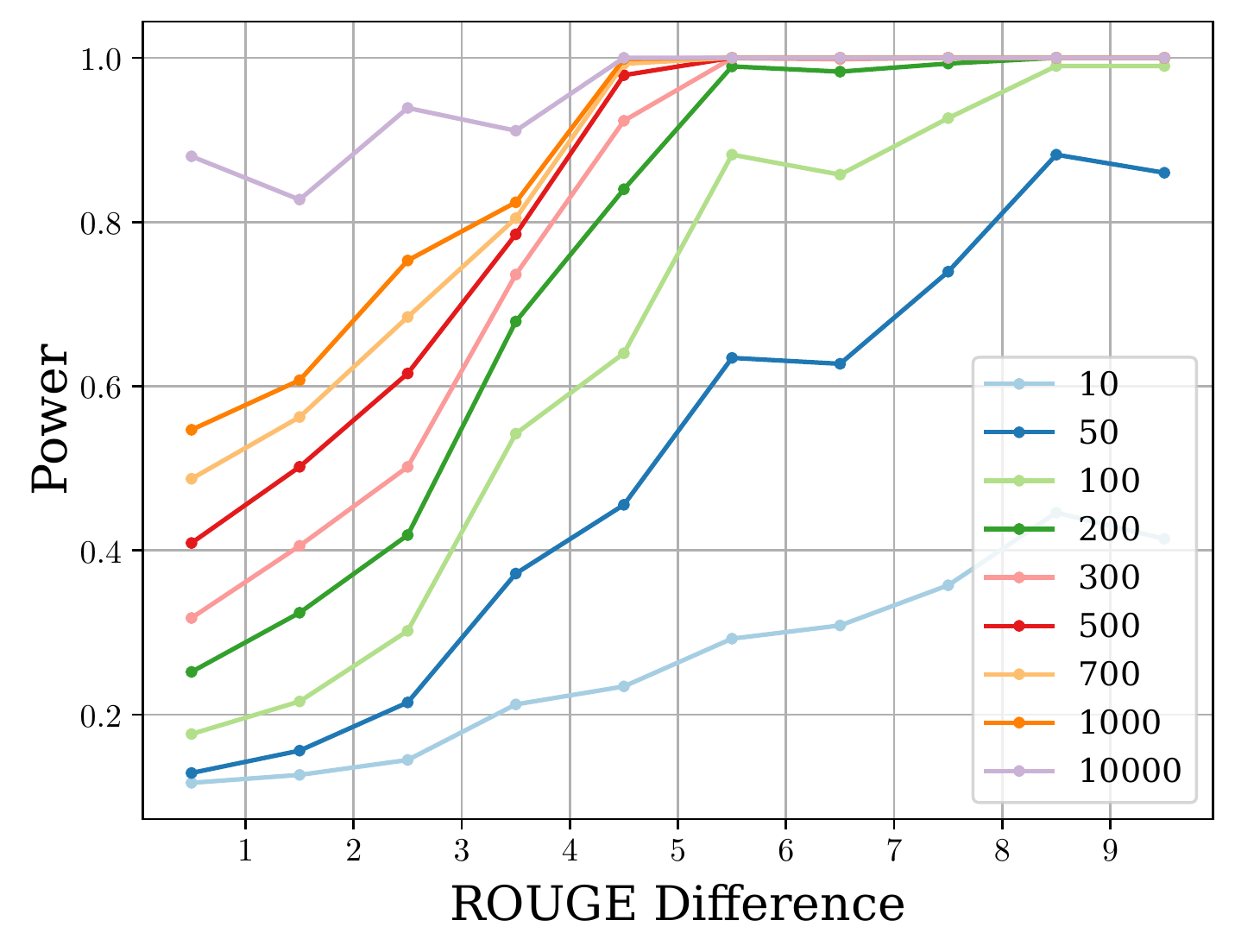}
 \caption{\textbf{Power analysis} of human evaluation for system comparison on the annotated CNNDM test examples. 
 Different lines represent results with different sample sizes.
 The system pairs are grouped by performance differences in ROUGE1 recall scores.}
 \label{fig:power-cnndm}
\end{figure}

We conduct the power analysis for \textit{pair-wise} system comparisons with ACU scores (Eq.~\ref{eq:acu}) focusing on two factors, the \textit{number of test examples} and the \textit{observed system difference}.
Specifically, we run the power analysis with varying sample sizes, and group the system pairs into buckets according to their performance difference, as determined by ROUGE1 recall scores (Fig.\ref{fig:power-cnndm}).\footnote{We note that these scores are proxies of the true system differences, and the power analysis is based on the assumption that the systems have significantly different performance.} 
We observe the following:
\noindent (1) \textbf{A high statistical power\footnote{An experiment is usually considered sufficiently powered if the statistical power is over 0.80.} is difficult to reach when the system performance is similar.}
Notably, while the sample size of the human evaluation performed in recent work is typically around 50-100,\footnote{We provide a brief survey of the practices of human evaluation in recent text summarization research in Appendix~\ref{appendix:survey}.} such sample size can only reach a power of 0.80 when the ROUGE1 recall score difference is above 5.
\noindent (2) \textbf{Increasing the sample size can effectively raise the statistical power. }
For example, when the system performance difference is within the range of 1-2 points, the power of a 500-sample set is around 0.50 while a 100-sample set only has a power of around 0.20.
The results of power analysis on three datasets with both ROUGE and ACU score differences are provided in Appendix~\ref{subsec:power-xsum-samsum} with the same patterns, which indicates that our dataset can provide more stable summarization system evaluation thanks to its higher statistical power.

\subsection{Summarization System Analysis}

\begin{table}[t!]
\small
\centering
\addtolength{\tabcolsep}{-1pt} 
\begin{tabular}{lcccc}
\toprule
 \textbf{System} &   \textbf{ACU} &   \textbf{\textit{n}ACU} &   \textbf{Len} &   \textbf{R1F} \\
\midrule
 \scriptsize GSum~\tiny\cite{dou-etal-2021-gsum}     &       44.47 &                  34.87 &    77.61 &   45.47 \\
 \scriptsize MatchSum~\tiny\cite{zhong-etal-2020-extractive} &       42.50  &                  33.69 &    74.99 &   43.84 \\
 \scriptsize BRIO-Ext~\tiny\cite{liu-etal-2022-brio} &       41.72 &                  33.58 &    73.67 &   44.44 \\
 \scriptsize BART~\tiny\cite{lewis-etal-2020-bart}     &       38.83 &                  32.34 &    71.00    &   44.04 \\
 \scriptsize CTRLSum~\tiny\cite{https://doi.org/10.48550/arxiv.2012.04281}  &       \textbf{44.58} &                  36.13 &    70.56 &   45.69 \\
 \scriptsize BRIO~\tiny\cite{liu-etal-2022-brio}     &       44.03 &                  \textbf{37.20}  &    69.58 &   \textbf{47.83} \\
 \scriptsize CLIFF~\tiny\cite{cao-wang-2021-cliff}    &       38.51 &                  32.96 &    67.74 &   44.19 \\
 \scriptsize PEGASUS~\tiny\cite{10.5555/3524938.3525989}  &       37.56 &                  32.03 &    65.65 &   43.80  \\
 \scriptsize SimCLS~\tiny\cite{liu-liu-2021-simcls}   &       40.47 &                  36.01 &    62.91 &   46.46 \\
 \scriptsize FROST~\tiny\cite{10.1162/tacl_a_00438}    &       38.44 &                  33.68 &    62.65 &   44.90  \\
 \scriptsize GOLD~\tiny\cite{pang2021text}     &       38.10  &                  33.80  &    60.65 &   44.86 \\
 \scriptsize GLOBAL~\tiny\cite{NEURIPS2021_89d4402d}     &       36.40  &                  34.07 &    55.50  &   45.17 \\
\bottomrule
\end{tabular}
\addtolength{\tabcolsep}{1pt} 
 % \vspace{-2mm}
\caption{Summarization system analysis on CNNDM.
\textbf{ACU} is the ACU score (Eq.~\ref{eq:acu}), \textbf{\textit{n}ACU} is the normalized ACU score (Eq.~\ref{eq:normalzied-acu}), \textbf{Len} is the average summary length, and \textbf{R1F} is the ROUGE1 F1 score.
\textbf{ACU} and \textbf{\textit{n}ACU} are calculated on the 500 annotated examples (the value is multiplied by 100) while \textbf{Len} and \textbf{R1F} are calculated on the entire test set. 
The systems are sorted by \textbf{Len}.
}
\label{tab:systems} 
\end{table}

\label{subsec:sys-analysis}

As a case study, in Tab.~\ref{tab:systems} we analyze the summary characteristics of the recent summarization systems we collected on the CNNDM test set.
XSum and SamSum results are shown in Appendix~\ref{subsec:append-sys-scores}.
Apart from the ACU scores, we note that \textbf{the average summary length of different systems can greatly vary}, and such differences are not always captured by the widely-used ROUGE F1.
For example, the length of GSum~\cite{dou-etal-2021-gsum} is around 40\% longer than GLOBAL~\cite{NEURIPS2021_89d4402d} while they have very similar ROUGE1 F1 scores.
Besides, we note \textbf{all systems in Tab.~\ref{tab:systems} have longer summaries than the reference summaries}, whose average length is only 54.93.
This can be a potential risk to users who may prefer shorter, more concise summaries.
Meanwhile, the systems that generate longer summaries may be favored by users who prefer more informative summaries. 
Therefore, we join the previous work~\cite{sun-etal-2019-compare, song-etal-2021-new, Gehrmann2022RepairingTC, goyal-gpt3} in advocating \textbf{treating summary lengths as a separate aspect of summary quality in evaluation}, as in earlier work in summarization research.\footnote{For example, the DUC evaluation campaigns set a pre-specified maximum summary length, or summary budget.} 

\newcommand{\preference}{Prior\xspace}
\newcommand{\salience}{Ref-free\xspace}
\newcommand{\accuracy}{Ref-based\xspace}

\section{Evaluating Annotation Protocols}
\label{sec:protocol-comparison}
Apart from ACU annotations, we collect human annotations with three different protocols to better understand their characteristics. 
Specifically, two reference-free protocols are investigated: \textit{\preference} protocol evaluates the annotators' preferences of summaries \textit{without} the input document, while \textit{\salience} protocol evaluates if summaries cover the salient information of the input document. 
We also consider one reference-based protocol, \textit{\accuracy}, which evaluates the content similarity between the generated and reference summaries. 
Appendix~\ref{subsec:protocol-details-appendix} provides detailed instructions for each protocol.

\subsection{Annotation Collection}
We collected three annotations per summary on a 100-example subset of the above CNNDM test set using the same pool of workers from our ACU qualification. 
Except for ACU, all of the summaries from different systems are evaluated within a single task with a score from 1 (worst) to 5 (best), similar to the EASL protocol~\citep{sakaguchi-van-durme-2018-efficient}.
We collect (1) annotations of the 12 above systems, with an inter-annotator agreement (Krippendorff's alpha) of 0.3455, 0.2201, 0.2741 on \textit{\preference}, \textit{\salience}, \textit{\accuracy} protocols respectively;
(2) annotations for summaries from GPT-3~\cite{NEURIPS2020_1457c0d6},\footnote{We use the ``text-davinci-002'' version of GPT-3.} T0~\cite{sanh2022multitask}, BRIO, and BART to better understand annotation protocols with respect to recently introduced large language models applied to zero-shot summarization.

\subsection{Results Analysis}
\label{subsec:protocal-result-analysis}

We investigate both the summary-level and system-level correlations of evaluation results of different protocols to study their inherent similarity.
Details of correlation calculation are in Appendix~\ref{appendix:corr}.

\begin{table}[t!]
\small
\centering
\addtolength{\tabcolsep}{-1.5pt} 
\begin{tabular}{lcccc}
\toprule
                &  \textbf{\preference}  & \textbf{\salience}  &  \textbf{\accuracy}  & \textbf{\textit{n}ACU}   \\
\midrule
 \preference    & -          & \textbf{0.926}      & -0.061      & 0.048                \\
 \salience       & 0.926      & -          & \textbf{-0.247}      & -0.093             \\
 \accuracy      & -0.061     & -0.247     & -           & 0.762              \\
 \textit{n}ACU & 0.048      & -0.093     & 0.762       & -                 \\
\midrule
 Len.        & 0.833      & 0.875      & -0.550      & -0.296              \\
\bottomrule
\end{tabular}
\addtolength{\tabcolsep}{1.5pt} 
\caption{\textit{System-level} Pearson's correlations between different protocols on the fine-tuned models. 
\textbf{\textit{n}ACU} is the normalized ACU score.
\textbf{Len.} is the summary length.
}
\label{tab:protocol_corr_sys} 
\end{table}

\noindent \textbf{Results on Fine-tuned Models}
We show the system-level protocol correlation when evaluating the fine-tuned models in Tab.~\ref{tab:protocol_corr_sys}, and the summary-level correlation can be found in Appendix~\ref{subsec:protocol-analysis-appendix}.
We use the \textit{normalized} ACU score (Eq.~\ref{eq:normalzied-acu}) because the other evaluation protocols are supposed to resemble an F1 score, while the ACU score is by definition recall-based.
We have the following observations:
\par \noindent (1) The \textit{\salience} protocol has a strong correlation with the \textit{\preference} protocol, suggesting that the latter may have a large impact on the annotator's document-based judgments.
\par \noindent (2) Both the \textit{\preference} and \textit{\salience} protocols have a strong correlation with summary length, showing that annotators may favor longer summaries.
\par \noindent (3) The \textit{\salience} protocol and the \textit{\accuracy} protocol have a negative correlation while ideally they are supposed to measure similar quality aspects.

We perform power analysis on the results following the procedure in \S\ref{subsec:power-analysis} and found that ACU protocol can yield higher statistical power than the \textit{\accuracy} protocol, suggesting that the ACU protocol leads to more robust evaluation results.
We also found that the reference-free \textit{\preference} and \textit{\salience} protocols have higher power than the reference-based protocols.
However, we note that they are not directly comparable because they have different underlying evaluation targets, as shown by the near-zero correlation between them.
Further details are provided in Appendix~\ref{subsec:protocol-analysis-appendix}.

\noindent \textbf{Results on Large Language Models}
The results are shown in Tab.~\ref{tab:llm}.
Apart from the system outputs, we also annotate reference summaries for reference-free protocols. 
We found that \textbf{under the \textit{\salience} protocol, GPT-3 receives the highest score while the reference summary is the least favorite one}, similar to the findings of recent work~\cite{goyal-gpt3, liang-hollistic}.
However, we found the same pattern with the \textit{\preference} protocol, showing that the \textbf{annotators have a \textit{prior} preference for GPT-3}. 
We provide an example in Appendix~\ref{subsec:protocol-analysis-appendix} comparing GPT-3 and BRIO summaries under different protocols.
Given the strong correlation between the \textit{\preference} and \textit{\salience} protocols, we note that there is a risk that the annotators' decisions are affected by their prior preferences that are not genuinely related to the task requirement. 
As a further investigation, we conduct an annotator-based case study including 4 annotators who annotated around 20 examples in this task, in which we compare two summary-level correlations (Eq. \ref{eq:summ_corr}) given a specific annotator: 
(1) the correlation between their own \textit{\salience} protocol scores and \textit{\preference} scores; 
(2) the correlation between their \textit{\salience} scores and the \textit{\salience} scores averaged over the other annotations on each example. 
We found that the average value of the former is 0.404 while the latter is only 0.188, suggesting that \textbf{the annotators' own \textit{\preference} score is a better prediction of their \textit{\salience} score than the \textit{\salience} score of other annotators}.

\begin{table}[t!]
\small
\centering
\addtolength{\tabcolsep}{-1pt} 
\begin{tabular}{lccccc}
\toprule
                &  \textbf{\preference}  & \textbf{\salience}  &  \textbf{\accuracy}  & \textbf{ACU} & \textbf{Len.}  \\
\midrule
 BART    & 3.58          & 3.52      & 2.93      & 0.367    &     69.5       \\
 BRIO       & 3.51      & 3.49         & \textbf{3.07}     & \textbf{0.429}    &    66.4     \\
 T0 & 3.33      & 3.24     &  2.84      &     0.295         & 61.6  \\
 GPT-3      & \textbf{3.72}     & \textbf{3.76}     & 2.74         &    0.268     &  69.5    \\
\midrule
 Ref.        & 2.85      & 2.94      & -      & -     &    54.9     \\
\bottomrule
\end{tabular}
\addtolength{\tabcolsep}{1pt} 
\caption{Model performance under different annotation protocols. 
\textbf{Len.} is the summary length.
\textbf{Ref.} is the reference summary.
\textit{\preference},\textit{ \salience}, \textit{\accuracy} protocols have a score range from 1 to 5.
}
\label{tab:llm} 
\end{table}

\section{Evaluating Automatic Metrics}
\label{sec:metric-comparison}

We analyze several representative automatic metrics, with additional results in Appendix \ref{append:metric_analysis} on 50 automatic metric variants.
We focus the metric evaluation on ACU annotations because of two insights from \S\ref{sec:protocol-comparison}:
(1) Reference-based metrics should be evaluated with reference-based human evaluation.
(2) ACU protocol provides higher statistical power than the summary-level \textit{\accuracy} protocol.

\subsection{Metric Evaluation with ACU Annotations}

\label{subsec:metric-eval}

\begin{table}[t!]
\small
\centering
\addtolength{\tabcolsep}{-2pt} 
\begin{tabular}{@{\extracolsep{1pt}}lcccccc@{}}
\toprule
 & \multicolumn{2}{c}{\textbf{CNNDM}} & \multicolumn{2}{c}{\textbf{XSum}} & \multicolumn{2}{c}{\textbf{SamSum}} \\
 &  \textbf{Sys.} & \textbf{Sum.} & \textbf{Sys.} & \textbf{Sum.} & \textbf{Sys.} & \textbf{Sum.} \\
 \cmidrule{2-3} \cmidrule{4-5} \cmidrule{6-7} 
ROUGE1                   & .788           & \textbf{.468}         & \textbf{.714}          & \textbf{.293}        & .929            & .439          \\
ROUGE2                   & .758           & .453         & .643          & .266        & \textbf{1.00}           & .395          \\
ROUGEL                  & \textbf{.879}           & .454         & .643          & .258        & .929            & .415          \\
 METEOR                    & .758           & .407         & .571          & .268        & .857            & .373          \\
CHRF                      & .758           & .436         & .571          & .275        & .857            & .396          \\
 BERTScore      & .515           & .448         & .571          & .277        & .857            & .417          \\
 BARTScore & .727           & .453         & \textbf{.714}          & .282        & .929            & .430          \\
 QAEval                & .849           & .358         & .429          & .198        & .929            & .384          \\
 SummaQA          & .727           & .119         & .143          & .019        & .643            & .102          \\
 Lite$^3$Pyramid                       & .849           & .452         & \textbf{.714}          & .245        & \textbf{1.00}           & \textbf{.467} \\   
 GPTScore         & .636           & .129         & .214          & .099        & .429            & .158          \\
 G-Eval-3.5            & .412           & .164         & .429          & .136        & .857            & .248          \\
G-Eval-3.5-S           & .364           & .171         & .429          & .144        & .857            & .262  \\
 G-Eval-4               & .779           & .274         & .691          & .185        & .929            & .405          \\
 \bottomrule
\end{tabular}
\addtolength{\tabcolsep}{2pt} 
\caption{The Kendall's correlation between the automatic metric scores and ACU scores of system outputs on CNNDM, XSum, and SamSum datasets.
The correlation is calculated at both the system level and the summary level. 
We use the \textit{recall} score of the automatic metrics when available to align with the ACU scores.
}
\label{tab:metric-corr} 
\end{table}

We use the correlations between automatic metric scores and ACU annotation scores of system outputs to analyze and compare automatic metric performance.
The following metrics are evaluated: 

(1) lexical overlap based metrics, \textbf{ROUGE}~\citep{lin-2004-rouge}, \textbf{METEOR}~\citep{lavie-agarwal-2007-meteor}, \textbf{CHRF}~\citep{popovic-2015-chrf}; 
(2) pre-trained language model based metrics, \textbf{BERTScore}~\citep{bert-score}, \textbf{BARTScore}~\citep{BARTScore_NEURIPS2021_e4d2b6e6}; 
(3) question-answering based metrics, \textbf{SummaQA}~\citep{scialom-etal-2019-answers}, \textbf{QAEval}~\citep{deutsch-etal-2021-towards}; 
(4) \textbf{Lite$^3$Pyramid}~\citep{zhang-bansal-2021-finding}, which automates the LitePyramid evaluation process;
(5) evaluation methods based on large language models, \textbf{GPTScore}~\cite{Fu2023GPTScoreEA} and \textbf{G-Eval}~\cite{Liu2023GEvalNE}, with two variants that are based on GPT-3.5\footnote{OpenAI's \texttt{gpt-3.5-turbo-0301}: \url{https://platform.openai.com/docs/models/gpt-3-5}.} (\textbf{G-Eval-3.5}) and GPT-4\footnote{OpenAI's \texttt{gpt-4-0314}: \url{https://platform.openai.com/docs/models/gpt-4}.}~\cite{OpenAI2023GPT4TR} (\textbf{G-Eval-4}) respectively.
We note that for LLM-based evaluation we require the metric to calculate the \textit{recall} score.
For G-Eval-3.5 we report two variants that are based on greedy decoding (G-Eval-3.5) and sampling (G-Eval-3.5-S) respectively,
Details of the LLM-based evaluation are in Appendix~\ref{appendix:llm-metric}.

Tab.~\ref{tab:metric-corr} shows the results, with additional results of more metrics in Appendix~\ref{subsec:appendix-metric-scores}.
We note:

\noindent (1) Several automatic metrics from the different families of methods (e.g., ROUGE, BARTScore) are all able to achieve a relatively high correlation with the ACU scores, especially at the system level.

\noindent (2) Metric performance varies across different datasets. 
In particular, metrics tend to have stronger correlations on the SamSum dataset and weaker correlations on the XSum dataset.
We hypothesize that one reason is that the reference summaries of the XSum dataset contain more complex structures.

\noindent (3) Despite their successes~\cite{Fu2023GPTScoreEA, Liu2023GEvalNE} in other human evaluation benchmarks such as SummEval, LLM-based automatic evaluation cannot outperform traditional methods such as ROUGE on RoSE.
Moreover, their low summary-level correlation with ACU scores suggests that their predicted scores may not be well-calibrated.

\begin{table}[t!]
\small
\centering
\addtolength{\tabcolsep}{-1.4pt} 
\begin{tabular}{lcccccc}
\toprule
 \textbf{Bucket} & \textbf{1} & \textbf{2} & \textbf{3} & \textbf{4} & \textbf{5} & \textbf{6} \\
 \midrule
ROUGE1                   & .091     & .636     & 1.00   & 1.00   & 1.00   & 1.00   \\
 ROUGE2                   & -.091    & .818     & .818     & 1.00   & 1.00   & 1.00   \\
 ROUGEL                   & .455     & .818     & 1.00   & 1.00   & 1.00   & 1.00   \\
 METEOR                    & .091     & .818     & .818     & .818     & 1.00   & 1.00   \\
 CHRF                      & .091     & .818     & .818     & .818     & 1.00   & 1.00   \\
 BERTScore       & -.091    & .636     & .636     & .091     & .818     & 1.00   \\
 BARTScore & -.091    & .818     & .818     & .818     & 1.00   & 1.00   \\
 QAEval                 & .455     & .818     & 1.00   & .818     & 1.00   & 1.00   \\
 SummaQA          & .636     & .818     & .636     & .273     & 1.00   & 1.00   \\
 Lite$^3$Pyramid                       & .273     & .818     & 1.00   & 1.00   & 1.00   & 1.00   \\
  GPTScore           & .273     & .091     & .455     & 1.00    & 1.00    & 1.00    \\
 G-Eval-3.5              & -.091    & -.273    & -.091    & .818     & 1.00    & 1.00    \\
 G-Eval-3.5-S &   -.091    & -.273    & -.273    & .818     & 1.00    & 1.00    \\
 G-Eval-4                 & .091     & .818     & .636     & 1.00    & 1.00   & 1.00    \\
 \bottomrule
\end{tabular}
\addtolength{\tabcolsep}{1.5pt} 
\caption{The \textit{system-level} Kendall's correlation between the automatic metric and ACU scores on different \textit{system pairs} grouped by their ACU score differences on the CNNDM dataset, into six equal-sized buckets.
We use the \textit{recall} score of the automatic metrics when available. 
}
\vspace{-4mm}
\label{tab:metric-corr-pairs} 
\end{table}

Following \citet{deutsch-etal-2022-examining}, we further investigate metric performance when evaluating system pairs with varying performance differences. 
Specifically, we group the system pairs based on the difference of their ACU scores into different buckets and calculate the modified Kendall's correlation~\citep{deutsch-etal-2022-examining} on each bucket.
The system pairs in each bucket are provided in Appendix~\ref{subse:system-pairs}.
Tab.~\ref{tab:metric-corr-pairs} shows that \textbf{the automatic metrics generally perform worse when they are used to evaluate similar-performing systems.}

\subsection{Analysis of Metric Evaluation}
\label{subsec:metric-analysis}

We analyze the metric evaluation with respect to the statistical characteristics and the impact of different human evaluation protocols on metric evaluation.

\noindent \textbf{Confidence Interval} We select several representative automatic metrics and calculate the confidence intervals of their system-level correlations with the ACU scores using bootstrapping. 
Similar to \citet{deutsch-etal-2021-statistical}, we find that the confidence intervals are large. 
However, we found that \textbf{having a larger sample size can effectively reduce the confidence interval}, which further shows the importance of increasing the \textit{statistical power} of the human evaluation dataset as discussed in \S\ref{subsec:power-analysis}.
We provide further details in Appendix~\ref{subsec:ci-appendix}.

\begin{figure}[t!]
\centering
\includegraphics[width=0.8\linewidth]{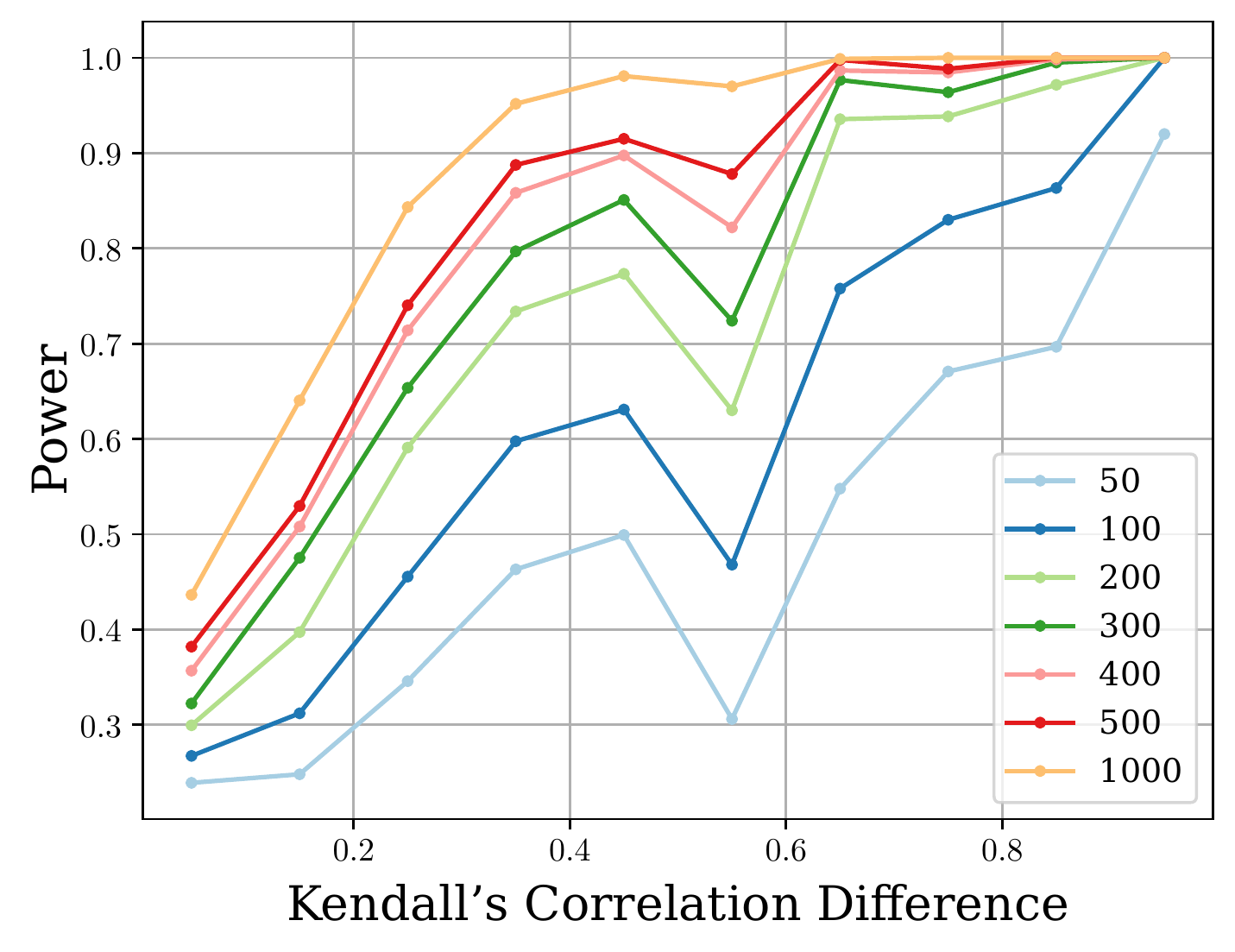}
% \vspace{-3mm}
 \caption{Power analysis of pair-wise metric comparison w.r.t. their \textit{system-level} Kendall's correlation coefficients with ACU scores on CNNDM.
 The metric pairs are grouped by the correlation differences with ACU scores.
 Different lines represent different sample sizes.
 }
 % \vspace{-7mm}
 \label{fig:metric-power}
\end{figure}

\noindent \textbf{Power Analysis of Metric Comparison}
We conduct a power analysis of \textit{pair-wise} metric comparison with around 200 pairs, which corresponds to the chance of a statistical significance result being found.
More details can be found in Appendix~\ref{subsec:power-analysis-metric-appendix}.
The results are in Fig.\ref{fig:metric-power}, showing similar patterns as in the power analysis of summarization system comparison (\S\ref{subsec:power-analysis}):

\noindent (1) Significant results are difficult to find when the metric performance is similar;

\noindent (2) Increasing the sample size can effectively increase the chance of finding significant results.

\begin{table}[t!]
\small
\centering
\addtolength{\tabcolsep}{-1pt} 
\begin{tabular}{lrrrr}
\toprule
 \textbf{Protocol} & \textbf{\preference} & \textbf{\salience} & \textbf{\accuracy} & \textbf{\textit{n}ACU}\\
 \midrule
ROUGE1                & -0.061        & -0.212      & 0.840       & 0.636           \\
 ROUGE2                & 0.000         & -0.151      & 0.595       & 0.636           \\
 ROUGEl                & -0.061        & -0.212      & 0.779       & 0.636           \\
 METEOR                & 0.394         & 0.242       & 0.382       & 0.485           \\
 CHRF                  & 0.576         & 0.424       & 0.199       & 0.485           \\
 BERTScore  & -0.091        & -0.182      & 0.779       & 0.485           \\
 BARTScore & -0.091        & -0.182      & 0.656       & 0.364           \\
 QAEval             & 0.485         & 0.515       & -0.076      & 0.151           \\
 SummaQA      & 0.515         & 0.424       & 0.260       & 0.303           \\
 Lite$^3$Pyramid                   & 0.576         & 0.667       & -0.168      & 0.121           \\
 \bottomrule
\end{tabular}
\addtolength{\tabcolsep}{1pt} 
\caption{The \textit{system-level} Kendall's correlation between the automatic metric and different human evaluation protocols on CNNDM dataset.
We use the \textit{F1} score of the automatic metrics when available. 
}
\label{tab:metric-corr-protocols} 
\end{table}

\noindent \textbf{Correlations under Different Human Evaluation Protocols}
We analyze the metric correlations under different human evaluation protocols (\S\ref{sec:protocol-comparison}).
The results are shown in Tab.~\ref{tab:metric-corr-protocols}, with more results in Appendix~\ref{subsec:corr-protocol-appendix}.
We note: 
(1) Metric performance differs greatly under different protocols, likely because the protocols can have weak correlations with each other (\S\ref{subsec:protocal-result-analysis}).
(2) The reference-based automatic metrics generally perform better under reference-based evaluation protocols, but can have negative correlations with reference-free protocols. 
\section{Conclusion and Implications}
\label{sec:concludsion-new}
We introduce \benchmark, a benchmark whose underlying protocol and scale allow for more robust summarization evaluation across three datasets. 
With our benchmark, we re-evaluate the current state of human evaluation and its implications for both summarization system and automatic metric development, and we suggest the following:

\noindent (1) \textbf{Alignment in metric evaluation}. To evaluate automatic metrics, it is important to use an appropriate human evaluation protocol that captures the intended quality dimension to be measured. 
For example, \textit{reference-based} automatic metrics should be evaluated by \textit{reference-based} human evaluation, which disentangles metric performance from the impact of reference summaries. 

\noindent (2) \textbf{Alignment in system evaluation}. We advocate for \textbf{\textit{targeted evaluation}}, which clearly defines the intended evaluation quality. 
Specifically, text summarization, as a conditional generation task, should be defined by both the source and target texts along with pre-specified, desired characteristics. 
Clearly specifying characteristics to be measured can lead to more reliable and objective evaluation results. This will be even more important for LLMs pre-trained with human preference feedback for disentangling annotators' \textit{prior} preferences for LLMs with the \textit{task-specific} summary quality. 

\noindent (3) \textbf{Alignment between NLP datasets and tasks}. 
Human judgments for summary quality can be diverse and affected by various factors such as summary lengths, and reference summaries are not always favored.
Therefore, existing summarization datasets (e.g. CNNDM) should \textit{only} be used for the appropriate tasks.
For example, they can be used to define a summarization task with specific requirements (e.g. maximum summary lengths), and be important for studying reference-based metrics.

\section{Limitations}\label{sec:limitations}
Biases may be present in the data annotator as well as in the data the models were pretrained on. 
Furthermore, we only include English-language data in our benchmark and analysis.
Recent work has noted that language models may be susceptible to learning such data biases \cite{gpt3-bias}, thus we request that the users be aware of potential issues in downstream use cases.
\par
As described in Appendix \ref{subsec:protocol-details-appendix}, we take measures to ensure a high quality benchmark. 
There will inevitably be noise in the dataset collection process, either in the ACU writing or matching step, and high agreement of annotations does not necessarily coincide with correctness. 
However, we believe that the steps taken to spot check ACU writing and filter workers for ACU matching allow us to curate a high-quality benchmark. 
Furthermore, we encourage the community to analyze and improve \benchmark in the spirit of evolving, living benchmarks \cite{gehrmann-etal-2021-gem}.
\par
For reference-based evaluation, questions about reference quality arise naturally.
We also note that the original Pyramid protocol was designed for multi-reference evaluation and weighting of semantic content units, while we do not weight ACUs during aggregation. 
As discussed above, we argue that our benchmark and analysis are still valuable given the purpose of studying conditional generation and evaluating automatic metrics for semantic overlap in targeted evaluation.
We view the collection of high-quality reference summaries as a valuable, orthogonal direction to this work, and we plan to explore ACU weighting in future work. 

\section*{Acknowledgements}
We thank the anonymous reviewers for their constructive comments. 
We are grateful to Arman Cohan for insightful discussions and suggestions, Daniel Deutsch for the initial discussions, Richard Yuanzhe Pang for sharing system outputs, and Philippe Laban for valuable comments.

% \clearpage
\bibliography{anthology,custom}
\bibliographystyle{acl_natbib}
\appendix
\section{Benchmark Data Collection}
\label{append:benchmark}
\subsection{Detailed Settings}
\label{append:settings}
We discuss the detailed settings of ACU collection in \S\ref{subsec:acu-protocol}.
To ensure the consistency of written ACUs among different annotators, we require each annotator to be familiar with the annotation protocol and proofread each other's annotations to resolve any differences in initial annotations.
After establishing a consistent understanding of the task, we have each reference summary annotated by one annotator. 
We note that there are multiple valid ways of writing the same atomic fact. 
In preliminary protocol analysis, we had multiple annotators write ACUs for the same reference summaries and did not find large differences in downstream inter-annotator agreement for ACU matching.
The average time to write ACUs of one summary ranges from 2 to 5 minutes, and the overall annotation time for ACU writing is around 150 hours.
\par 
We use the following qualifications, in addition to a qualification test, to recruit MTurk workers with good track records: 
HIT approval rate greater than or equal to 98\%, number of HITs approved greater than or equal to 10000, and located in either the United Kingdom or the United States.
 Workers were compensated between \$0.15 and \$0.55 per summary-level ACU HITs, with HITs bucketed according to the number of ACUs to be matched. 
 For protocol comparison HITs, workers were compensated between \$1 and \$3. 
 All HITs were carefully calibrated to equal a \$12/hour pay rate. 
 \par

The datasets we used for the collection are CNNDM, XSum and SamSum. 
The data release licenses are the Apache License for CNNDM and XSum, and CC BY-NC-ND 4.0 for SamSum.
Our collected benchmark will be released under the 3-Clause BSD license.

\subsection{Summarization Models}
\label{subsec:models-appendix}
We list the summarization models for ACUs annotations on CNNDM, XSum, and SamSum in \S\ref{sec:results}.

\paragraph{CNNDM Systems:}
\mbox{}
\par \noindent  \textbf{BART} \citep{lewis2019bart} introduce  a denoising autoencoder for pretraining sequence to sequence tasks which is applicable to both natural language understanding and generation tasks.
\par \noindent  \textbf{Pegasus} \citep{zhang2019pegasus}  introduce a model pretrained with a novel objective function designed for summarization by which important sentences are removed from an input document and then generated from the remaining sentences. 
\par \noindent  \textbf{MatchSum}~\cite{zhong-etal-2020-extractive} propose a summary-level extractive system using semantic match between the extracted summary and the source document. 
\par \noindent  \textbf{CTRLSum}~\cite{https://doi.org/10.48550/arxiv.2012.04281} introduce a method for controllable summarization based on keyword or descriptive prompt control tokens.
\par \noindent  \textbf{CLIFF}~\cite{cao-wang-2021-cliff} propose to use contrastive learning to improve factual consistency. We use the CLIFF output that uses an underlying BART model.
\par \noindent  \textbf{GOLD}~\cite{pang2021text} frames text generation as an offline reinforcement learning problem, using importance weighting and assigning weights to examples that receive a higher probability from the generation model.
\par \noindent  \textbf{GSum}~ \cite{dou-etal-2021-gsum} is a framework for incorporating forms of summarization guidance.
\par \noindent  \textbf{SimCLS}~\cite{liu-liu-2021-simcls} is a two-stage summarization model where candidates from BART are reranking by a RoBERTa \cite{liu2019roberta} scoring model trained using contrastive learning.

\par \noindent  \textbf{FROST}~\cite{10.1162/tacl_a_00438} propose to do content planning in both pretraining and finetuning summarization models with plans in the form of entity chains.   
\par \noindent  \textbf{GLOBAL}~\cite{NEURIPS2021_89d4402d} propose a variation of beam search that takes into account the global attention distribution.   

\par \noindent  \textbf{BRIO}~\cite{liu-etal-2022-brio}  proposes to train a summarization model both as a token-level generator and an evaluator of sequence candidates through contrastive reranking. 
\par \noindent  \textbf{BRIO-Ext}~\cite{liu-etal-2022-brio} uses BRIO's reranker on candidate extractive summaries from MatchSum.

\par
\par \noindent  The following models were included in protocol-comparison annotations.
\par \noindent  \textbf{T0}~\cite{sanh2022multitask} introduces a prompt-based model that is fine-tuned on multiple tasks, including summarization.
\par \noindent  \textbf{GPT-3}~\cite{NEURIPS2020_1457c0d6} is the davinci-002 model trained on human demonstrations and model outputs highly rated by humans.\footnote{\url{https://beta.openai.com/docs/model-index-for-researchers}}
\mbox{}\\
\paragraph{XSum Systems}
\mbox{}
For XSum we reuse several of the above models with their XSum-trained checkpoints as well as several variations from the above paper due to the scarcity of widely-available, easily-reproducible XSum outputs.

\par \noindent  \textbf{BART} \citep{lewis2019bart}  
\par \noindent  \textbf{Pegasus} \citep{zhang2019pegasus}
\par \noindent  \textbf{CLIFF}~\cite{cao-wang-2021-cliff}
\par \noindent  \textbf{CLIFF-Pegasus}~\cite{cao-wang-2021-cliff} is the CLIFF algorithm applied with Pegasus as the underlying model. 
  \par \noindent  \textbf{FROST}~\cite{10.1162/tacl_a_00438} 
\par \noindent  \textbf{BRIO}~\cite{liu-etal-2022-brio}
\par \noindent  \textbf{BRIO-ranking}~\cite{liu-etal-2022-brio}  is the paper's reranking model.
\par \noindent  \textbf{BART-beam-patience} \citep{kasai2022beam}
\mbox{}\\
\paragraph{SamSum Systems}
\mbox{}\\
We use system outputs from \citet{gao-wan-2022-dialsummeval}.
\par \noindent  \textbf{BART} \citep{lewis2019bart}  
\par \noindent  \textbf{Pegasus} \citep{zhang2019pegasus}
\par \noindent  \textbf{UniLM} \citep{dong2019unified} is a model pretrained on unidirection, bidirection, and sequence-to-sequence language modeling tasks.
\par \noindent  \textbf{Ctrl-DiaSumm} \citep{liu-chen-2021-controllable} propose controlled generation using named entity plans.
\par \noindent  \textbf{PLM-BART} \citep{feng-etal-2021-language} use DialogGPT \cite{zhang-etal-2020-dialogpt} to annotate input dialogues before finetuning.
\par \noindent  \textbf{CODS} \citep{wu-etal-2021-controllable} propose a two-stage generation model that first generates a sketch that is then used as a signal to the second-stage summarizer.
\par \noindent  \textbf{MV-BART} \citep{chen-yang-2020-multi} propose a multi-view encoder and a decoder that attends to these conversation views.
\par \noindent  \textbf{S-BART} \citep{chen-yang-2020-multi} encodes utterances as well as action and discourse graphs and introduces a decoder that attends to these different levels of granularity.

\subsection{ACU Scores of Summarization Models}
\label{subsec:append-sys-scores}

\begin{table}[t!]
\small
\centering
\addtolength{\tabcolsep}{-1pt} 
\begin{tabular}{lcccc}
\toprule
 \textbf{System} &   \textbf{ACU} &   \textbf{\textit{n}ACU} &   \textbf{Len} &   \textbf{R1F} \\
\midrule
 \scriptsize PATIENCE~\tiny\cite{kasai2022beam} &       27.11 &                  26.59 &    25.00    &   45.07 \\
 \scriptsize CLIFF$_P$~\tiny\cite{cao-wang-2021-cliff}       &       25.13 &                  24.94 &    21.29 &   46.20  \\
 \scriptsize BRIO-Mul~\tiny\cite{liu-etal-2022-brio}               &       26.34 &                  26.15 &    21.15 &   48.73 \\
 \scriptsize BART~\tiny\cite{lewis-etal-2020-bart}               &       23.99 &                  23.78 &    20.98 &   45.56 \\
 \scriptsize PEGASUS~\tiny\cite{10.5555/3524938.3525989}            &       24.83 &                  24.67 &    20.21 &   46.84 \\
 \scriptsize CLIFF$_B$~\tiny\cite{cao-wang-2021-cliff}               &       22.09 &                  21.93 &    20.17 &   44.52 \\
 \scriptsize FROST~\tiny\cite{10.1162/tacl_a_00438}   &       27.93 &                  27.77 &    19.86 &   47.83 \\
 \scriptsize BRIO-Ctr~\tiny\cite{liu-etal-2022-brio}      &       26.42 &                  26.29 &    19.65 &   48.06 \\
\bottomrule
\end{tabular}
\addtolength{\tabcolsep}{1pt} 
\caption{Summarization system analysis on the XSum dataset.
\textbf{ACU} is the ACU score (Eq.~\ref{eq:acu}), \textbf{\textit{n}ACU} is the normalized ACU score (Eq.~\ref{eq:normalzied-acu}), \textbf{Len} is the average summary length, and \textbf{R1F} is the ROUGE1 F1 score.
\textbf{ACU} and \textbf{\textit{n}ACU} are calculated on the 500 annotated examples (the value is multiplied by 100) while \textbf{Len} and \textbf{R1F} are calculated on the entire test set. 
The systems are sorted by \textbf{Len}.
CLIFF$_P$ is based on PEGASUS, while CLIFF$_B$ is based on BART.
}
\label{tab:systems-xsum-appendix} 
\end{table}

\begin{table}[t]
\small
\centering
\addtolength{\tabcolsep}{-2.5pt} 
\begin{tabular}{lcccc}
\toprule
 \textbf{System} &   \textbf{ACU} &   \textbf{\textit{n}ACU} &   \textbf{Len} &   \textbf{R1F} \\
\midrule
 \scriptsize MV-BART \tiny \citep{chen-yang-2020-multi}     &       47.65 &                  33.01 &    23.57 &   53.80  \\
 \scriptsize Ctrl-DiaSumm \tiny \citep{liu-chen-2021-controllable} &       49.05 &                  37.20  &    22.97 &   56.33 \\
 \scriptsize PLM-BART \tiny  \citep{feng-etal-2021-language}     &       43.74 &                  32.61 &    21.43 &   53.46 \\
 \scriptsize S-BART \tiny \citep{chen-yang-2020-multi}       &       34.57 &                  25.95 &    21.01 &   50.36 \\
 \scriptsize CODS  \tiny \citep{wu-etal-2021-controllable}        &       38.40  &                  33.41 &    20.11 &   52.48 \\
 \scriptsize BART  \tiny \citep{lewis2019bart}       &       42.85 &                  34.05 &    19.62 &   52.30  \\
 \scriptsize UniLM  \tiny \citep{dong2019unified}      &       32.74 &                  26.10  &    18.84 &   49.20  \\
 \scriptsize PEGASUS  \tiny \citep{zhang2019pegasus}    &       37.02 &                  31.99 &    17.32 &   50.87 \\
\bottomrule
\end{tabular}
\addtolength{\tabcolsep}{2.5pt} 
\caption{Summarization system analysis on the SamSum dataset.
\textbf{ACU} is the ACU score (Eq.~\ref{eq:acu}), \textbf{\textit{n}ACU} is the normalized ACU score (Eq.~\ref{eq:normalzied-acu}), \textbf{Len} is the average Summary length, and \textbf{R1F} is the ROUGE1 F1 score.
\textbf{ACU} and \textbf{\textit{n}ACU} are calculated on the 500 annotated examples (the value is multiplied by 100) while \textbf{Len} and \textbf{R1F} are calculated on the entire test set. 
The systems are sorted by \textbf{Len}.
}
\label{tab:systems-samsum-appendix} 
\end{table}

We report the ACU scores of the summarization systems we annotated on the XSum and SamSum datasets (\S\ref{subsec:sys-analysis}) in Tab.~\ref{tab:systems-xsum-appendix} and Tab.~\ref{tab:systems-samsum-appendix}, respectively.
The results on CNNDM can be found in Tab.~\ref{tab:systems}.
For the normalized ACU score (Eq.~\ref{eq:normalzied-acu}), we set the normalization strength $\alpha$ to 2, 5, 0.5, on CNNDM, XSum, SamSum, respectively, by a grid search for de-correlating the summary length and the normalized score at the summary level.

\section{Power Analysis}
\label{append:power}

\subsection{Detailed Settings}
\label{subsec:power-details}
We describe the algorithm for the power analysis in \S\ref{subsec:power-analysis} in Alg.\ref{alg:power-analysis}.
While prior work~\cite{card-etal-2020-little, wei-jia-2021-statistical} uses parametric methods to estimate statistical power, we conduct the power analysis with the bootstrapping test~\cite{tibshirani1993introduction} as recent work~\cite{deutsch-etal-2021-statistical} has shown that the assumptions of the parametric methods do not always hold for human evaluation of text summarization. 
The process involves (1) iteratively sampling a set of examples with a certain sample size from an existing dataset, (2) running the significance test on the sampled set, and (3) estimating the power by averaging across the trials. 

The essence of the test is to have a series of \textit{simulated} datasets sampled from the existing dataset and run the significance test on the sampled sets.
Here the existing dataset $X$ consists of human-annotated scores of system outputs.
We use paired bootstrapping for the significance test.
The power analysis is conducted over all the system pairs.

\begin{algorithm}[t]
\caption{Power Analysis}\label{alg:power-analysis}
\begin{algorithmic}

\Require $n$ (Sample Size)
\Require $X$ (Existing Dataset)
\Require $m$ (Trial Number)
\Ensure $p$ (Statistical Power)
\State $p \gets 0$
\For{$i=0$ to $m$}

\State $\hat{X}\gets$ Sampling $n$ examples from $X$ with replacement
\State $\Tilde{p} \gets$ Running the significance test on $\hat{X}$
\If{$\Tilde{p} < 0.05$}
    \State $p \gets p + 1$
\EndIf
\EndFor
\State $p \gets p / m$
\Return $p$
\end{algorithmic}
\end{algorithm}

\subsection{Powers of ACU Annotations}
\label{subsec:power-xsum-samsum}

\begin{figure*}[t!]
    \centering
    \begin{subfigure}[b]{0.45\textwidth}
         \includegraphics[width=1\linewidth]{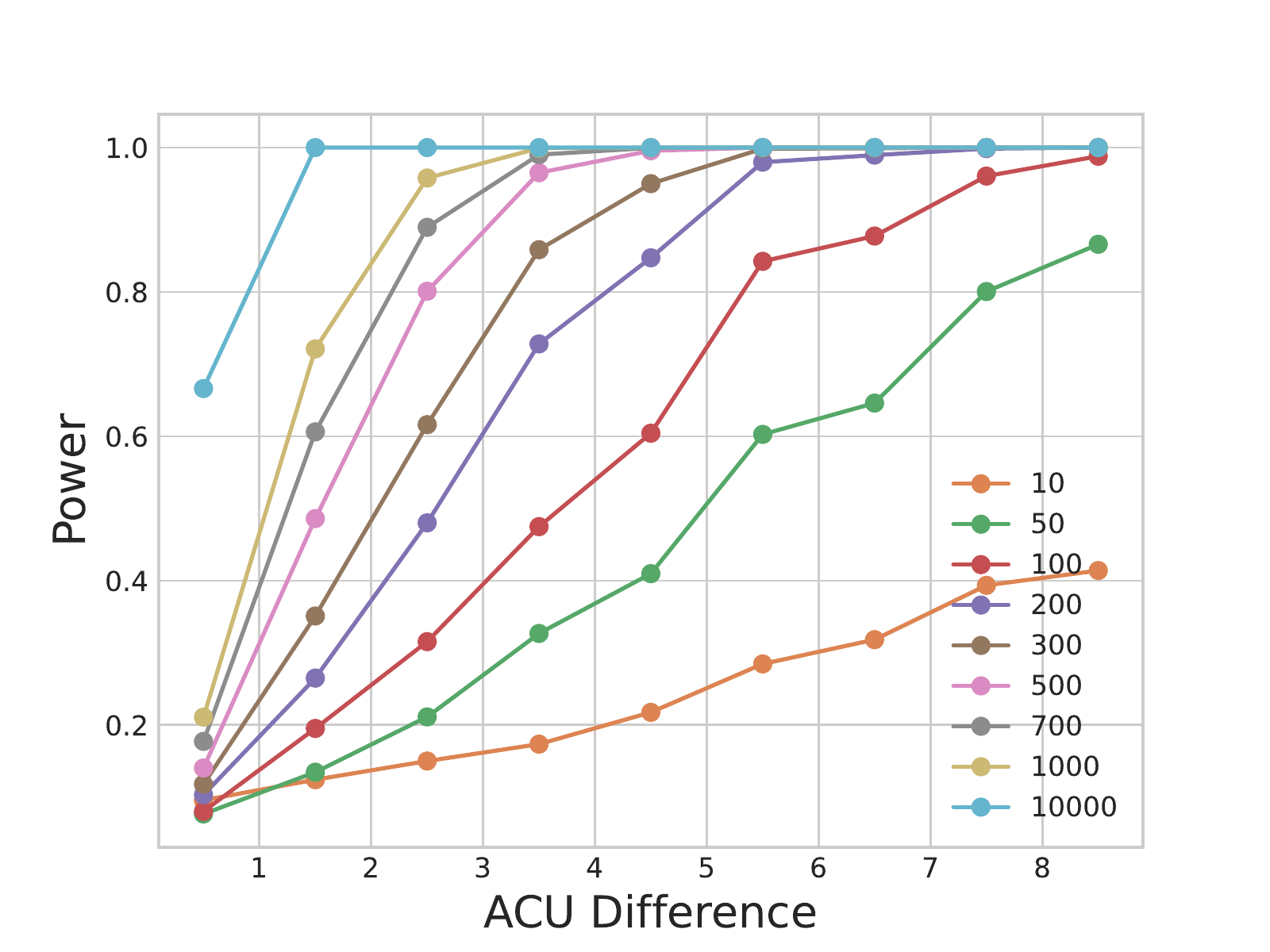}
\caption{\label{subfig:power-cnndm-human}}
    \end{subfigure}
\begin{subfigure}[b]{0.45\textwidth}
         \includegraphics[width=1\linewidth]{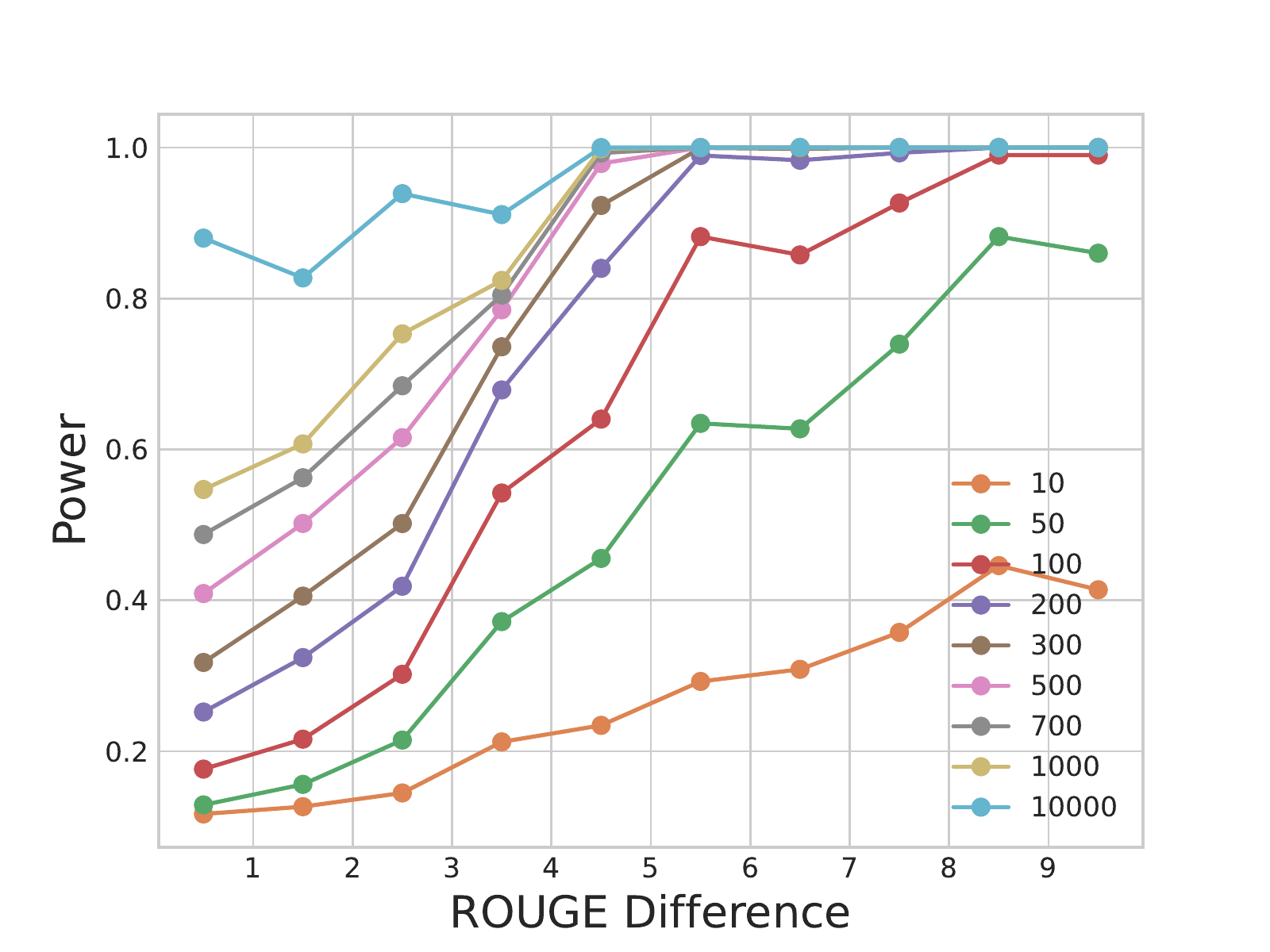}
\caption{\label{subfig:power-cnndm-ROUGE}}
    \end{subfigure}
 \caption{Power analysis of human evaluation for system comparison on the annotated CNNDM test examples. 
 Different lines represent results with different sample sizes.
 The system pairs are grouped by performance differences in ACU scores in Fig.\ref{subfig:power-cnndm-human}, and by ROUGE1 recall scores in Fig.\ref{subfig:power-cnndm-ROUGE}.
 \label{fig:power-cnndm-appendix}}
\end{figure*}

\begin{figure*}[t!]
    \centering
    \begin{subfigure}[b]{0.45\textwidth}
         \includegraphics[width=1\linewidth]{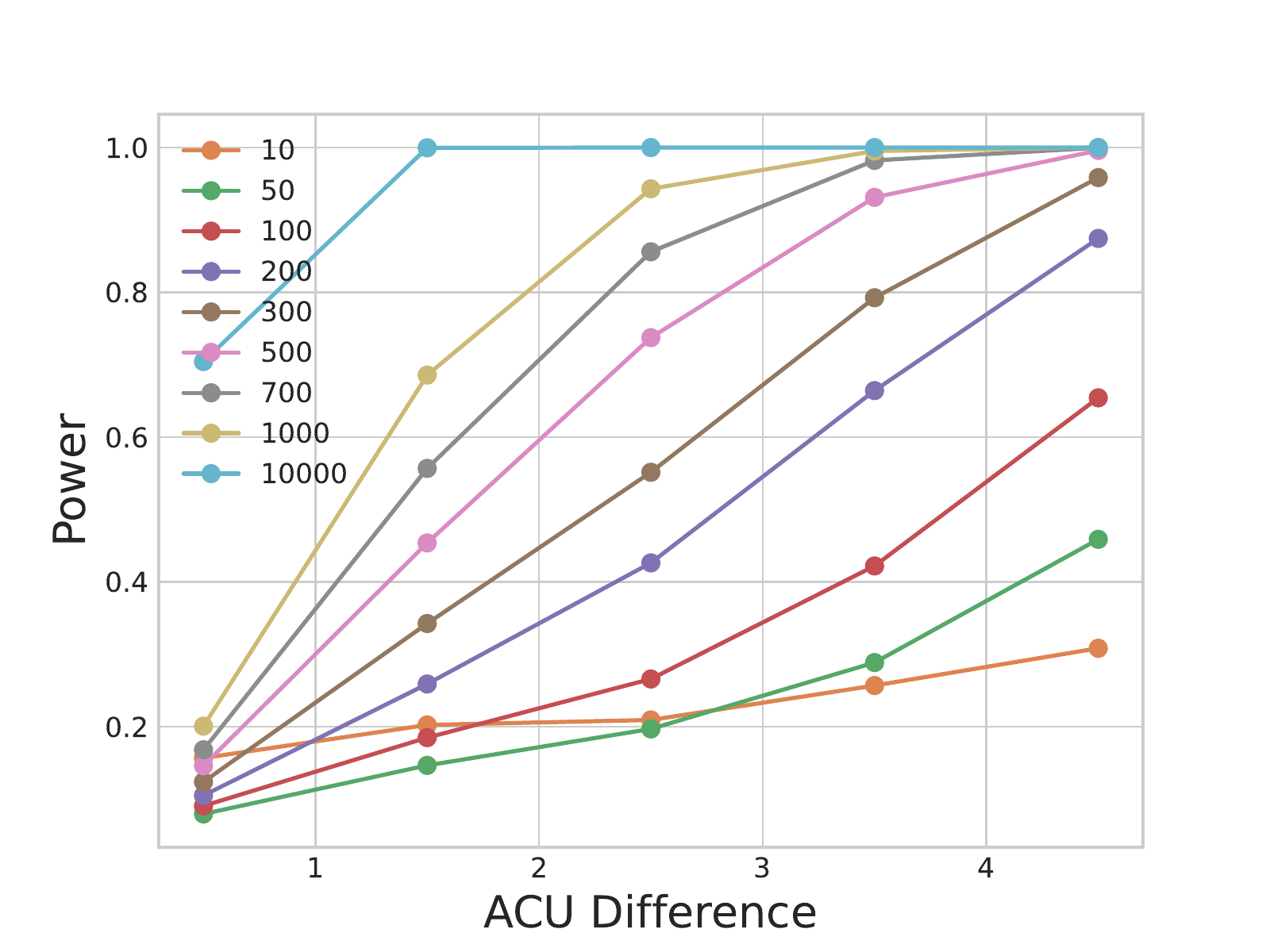}
\caption{\label{subfig:power-xsum-human}}
    \end{subfigure}
\begin{subfigure}[b]{0.45\textwidth}
         \includegraphics[width=1\linewidth]{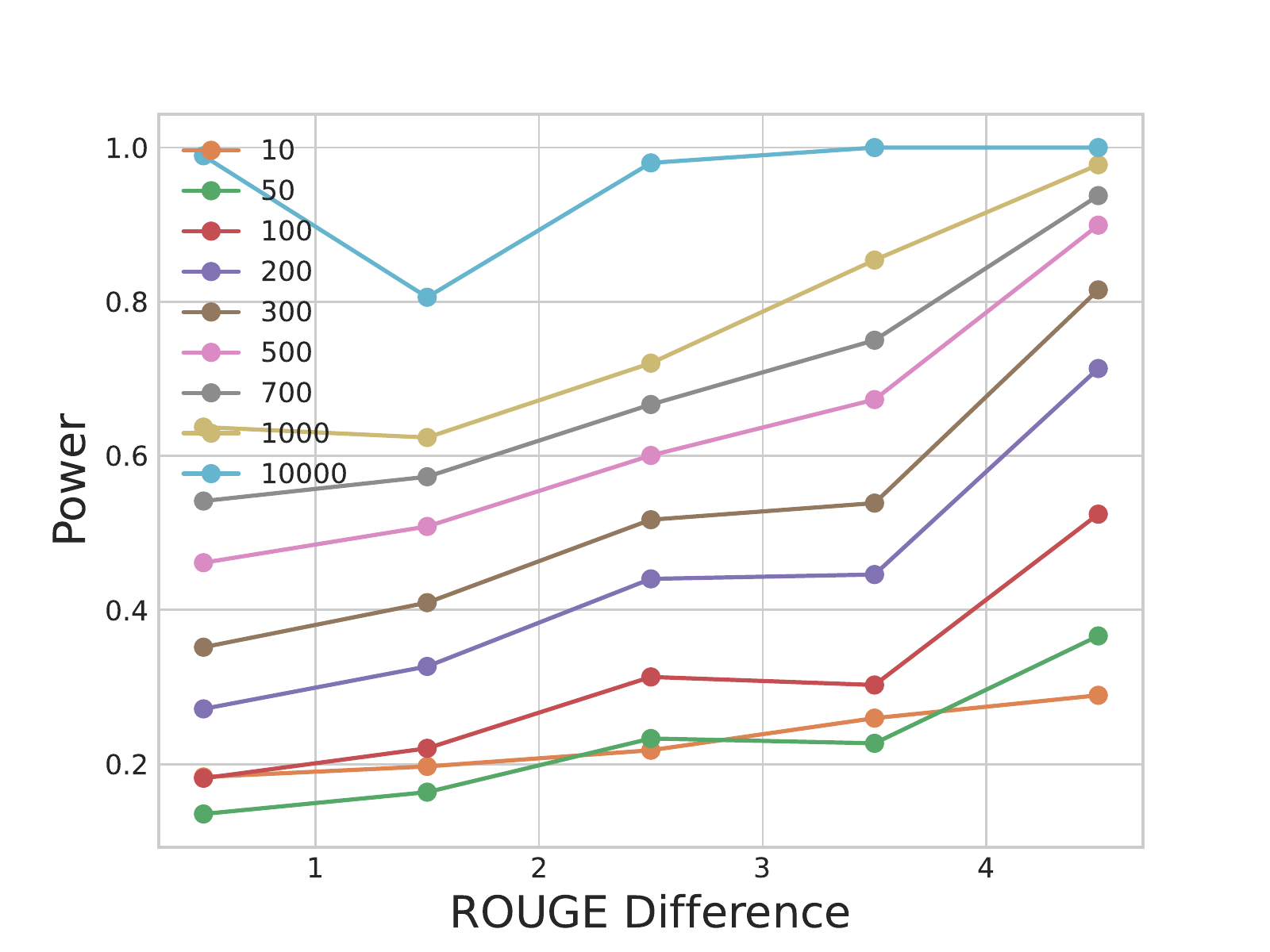}
\caption{\label{subfig:power-xsum-ROUGE}}
    \end{subfigure}
 \caption{Power analysis of human evaluation for system comparison on the annotated XSum test examples. 
 Different lines represent results with different sample sizes.
 The system pairs are grouped by performance differences in ACU scores in Fig.\ref{subfig:power-xsum-human}, and by ROUGE1 recall scores in Fig.\ref{subfig:power-xsum-ROUGE}.
 \label{fig:power-xsum}}
\end{figure*}

\begin{figure*}[t!]
    \centering
    \begin{subfigure}[b]{0.45\textwidth}
         \includegraphics[width=1\linewidth]{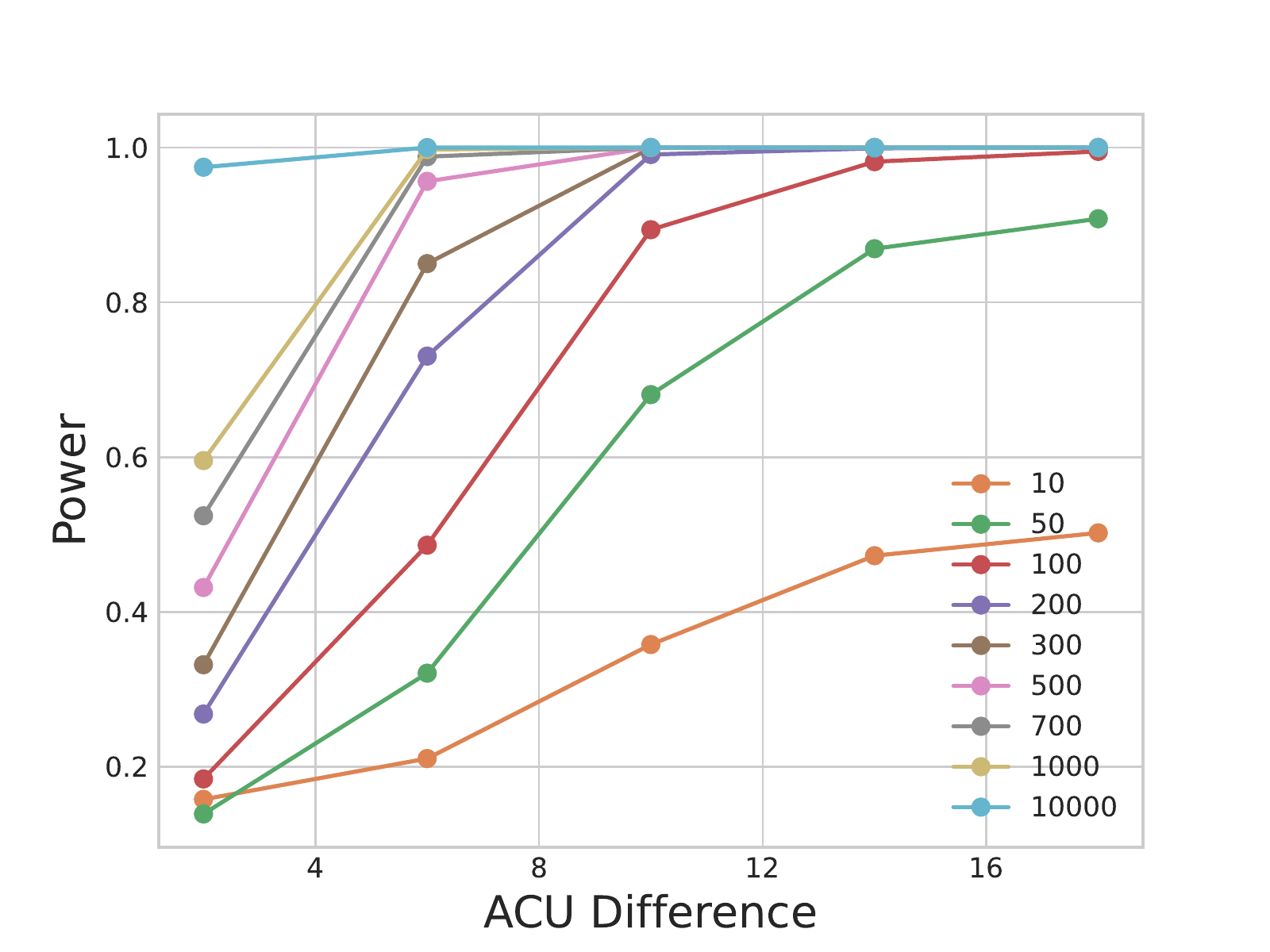}
\caption{\label{subfig:power-samsum-human}}
    \end{subfigure}
\begin{subfigure}[b]{0.45\textwidth}
         \includegraphics[width=1\linewidth]{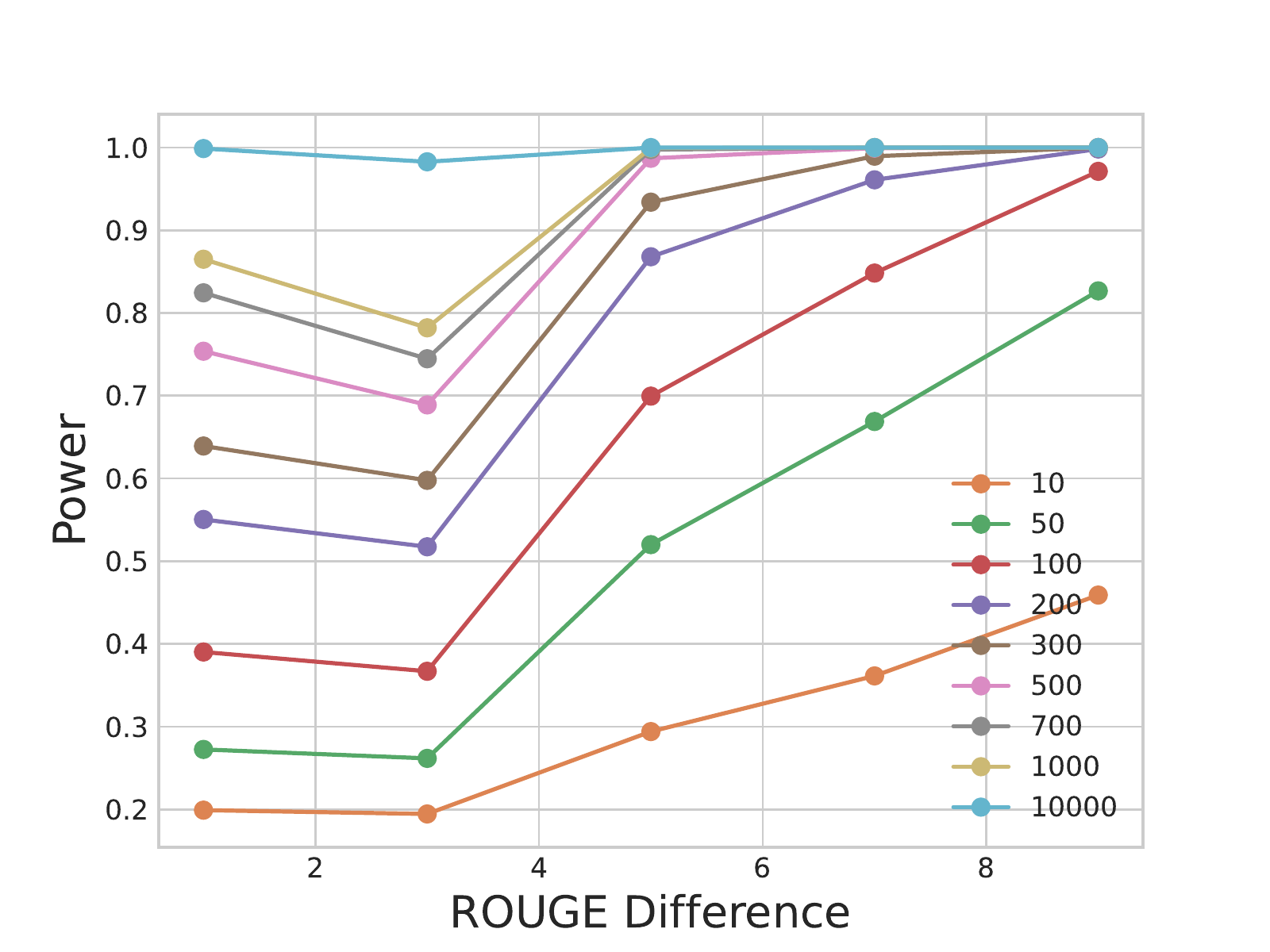}
\caption{\label{subfig:power-samsum-ROUGE}}
    \end{subfigure}
 \caption{Power analysis of human evaluation for system comparison on the annotated SamSum test examples. 
 Different lines represent results with different sample sizes.
 The system pairs are grouped by performance differences in ACU scores in Fig.\ref{subfig:power-samsum-human}, and by ROUGE1 recall scores in Fig.\ref{subfig:power-samsum-ROUGE}.
 \label{fig:power-samsum}}
\end{figure*}

Fig.\ref{fig:power-cnndm-appendix}, Fig.\ref{fig:power-xsum}, and Fig.\ref{fig:power-samsum} show the power analysis results on CNNDM, XSum and SamSum respectively in \S\ref{subsec:power-analysis}, where the system pairs are grouped by their performance difference in either ACU or ROUGE1 recall scores.
Similar to our findings on CNNDM in \S\ref{subsec:power-analysis}, we observe that increasing the sample size can effectively raise the statistical power.

\section{Calculating Correlations}
\label{appendix:corr}
We use correlations to analyze the inherent similarity between different human evaluation protocols, and the performance of automatic metrics, which is evaluated based on the correlations between the metric-calculated summary scores and the human-annotated summary scores.
Specifically, given $m$ system outputs on each of the $n$ data samples and two different evaluation methods (e.g., human evaluation and an automatic metric) resulting in two $n$-row, $m$-column score matrices $X$ and $Y$, the summary-level correlation is an average of sample-wise correlations:
\begin{equation}
\label{eq:summ_corr}
    r_{sum}(X, Y) = \frac{\sum_i \mathcal{C}(X_i, Y_i)}{n}, 
\end{equation}
where $X_i$, $Y_i$ are the evaluation results on the $i$-th data sample and $\mathcal{C}$ is a function calculating a correlation coefficient (e.g., the Pearson correlation coefficient).
In contrast, the system-level correlation is calculated on the aggregated system scores:
\begin{equation}
\label{eq:sys_corr}
    r_{sys}(X, Y) = \mathcal{C}(\bar{X}, \bar{Y}), 
\end{equation}
where $\bar{X}$ and $\bar{Y}$ contain $m$ entries which are the system scores from the two evaluation methods averaged across $n$ data samples, e.g., $\bar{X}_0 = \sum_i X_{i,0} / n$.

\section{Protocol Comparison}

\subsection{Data Collection Details}
\label{subsec:protocol-details-appendix}

\begin{table}[t!]
\small
\centering
\begin{tabular}{lccc}
\toprule
\textbf{Protocol} & \textbf{w/ Doc} & \textbf{w/ Ref} & \textbf{Fine-grained} \\
\midrule
\preference &  \xmark &  \xmark &   \xmark  \\
\salience  &  \cmark  &  \xmark  &   \xmark  \\
\accuracy &  \xmark  & \cmark  & \xmark   \\
ACU &  \xmark & \cmark & \cmark \\ 
\bottomrule
\end{tabular}
\caption{Human evaluation protocol comparison.
We categorize the different protocols based on if they (1) require the input document (\textbf{w/ Doc}), (2) require the reference summary (\textbf{w/ Ref}), and (3) are \textbf{fine-grained}.}
\label{tab:protocals-appendix} 
\end{table}

The 100 examples chosen for annotation in \S\ref{sec:protocol-comparison} are a subset of the CNNDM ACU test set, and as here we aim to analyze trends among protocols as opposed to observing statistically significant differences among systems, we believe 100 examples suffice for this collection.

We summarize and compare different protocols in Tab.~\ref{tab:protocals-appendix}.
We provide the following instructions to annotators for non-ACU annotations. We will release the full interface and instructions. 
\par \noindent \textbf{\preference}: We ask the annotator to imagine each of the candidate summaries to be evaluated as a summary of a longer news article and answer the following question: how good do you think this summary is?
\par \noindent \textbf{\salience}: The rating measures how well the summary captures the key points of the news article. Consider whether all and only the important aspects are contained in the summary.
\par \noindent \textbf{\accuracy}: The rating measures how similar two summaries are. The similarity depends on if the summaries contain similar information, not if they use the same words.

\subsection{Results Analysis}
\label{subsec:protocol-analysis-appendix}
We present the result analysis of \S\ref{subsec:protocal-result-analysis} here.
\begin{figure*}[t!]
\centering
\begin{subfigure}[b]{0.45\textwidth}
         \includegraphics[width=1\linewidth]{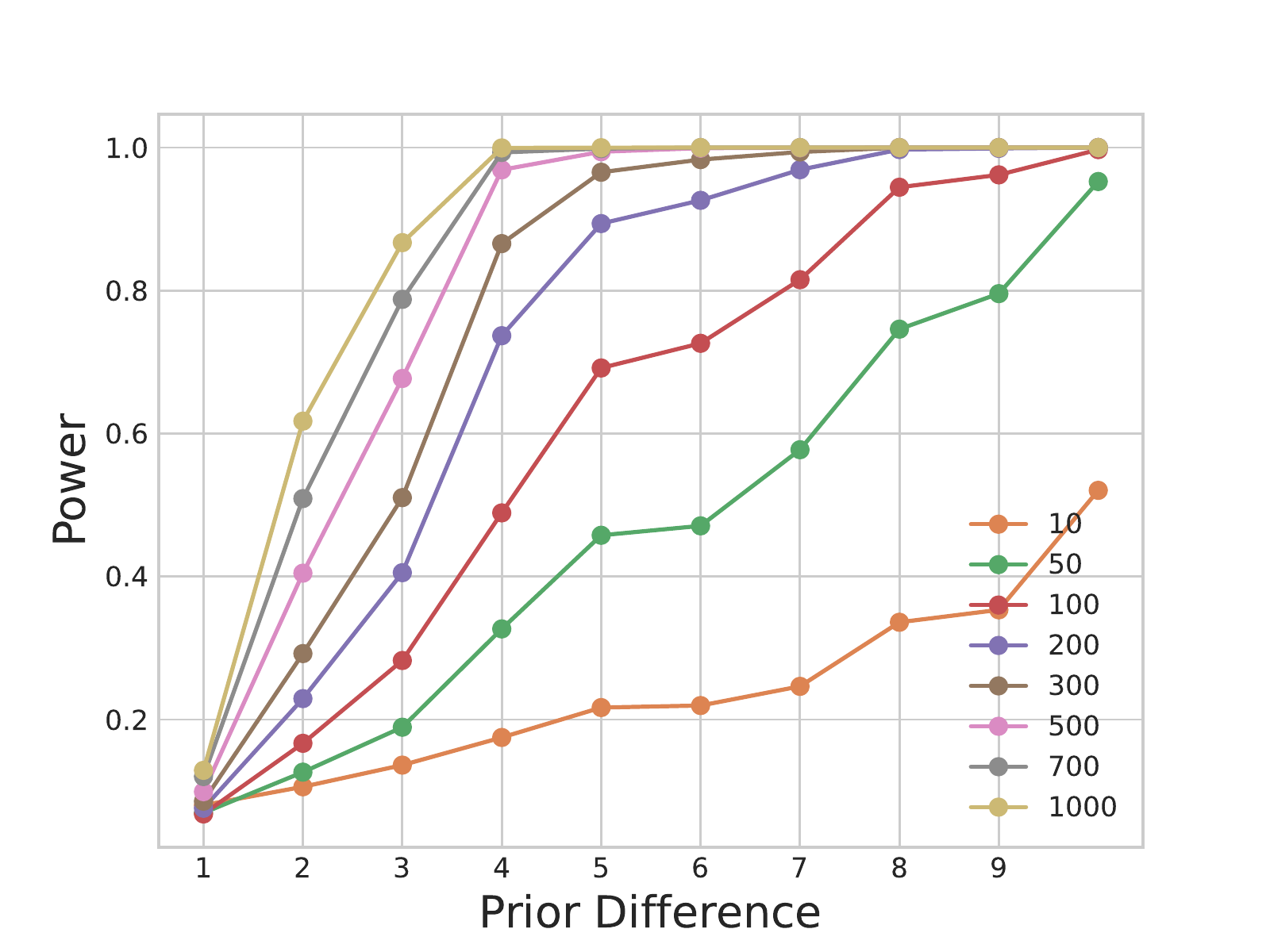}
\caption{\label{subfig:power-prior}}
    \end{subfigure}
\begin{subfigure}[b]{0.45\textwidth}
         \includegraphics[width=1\linewidth]{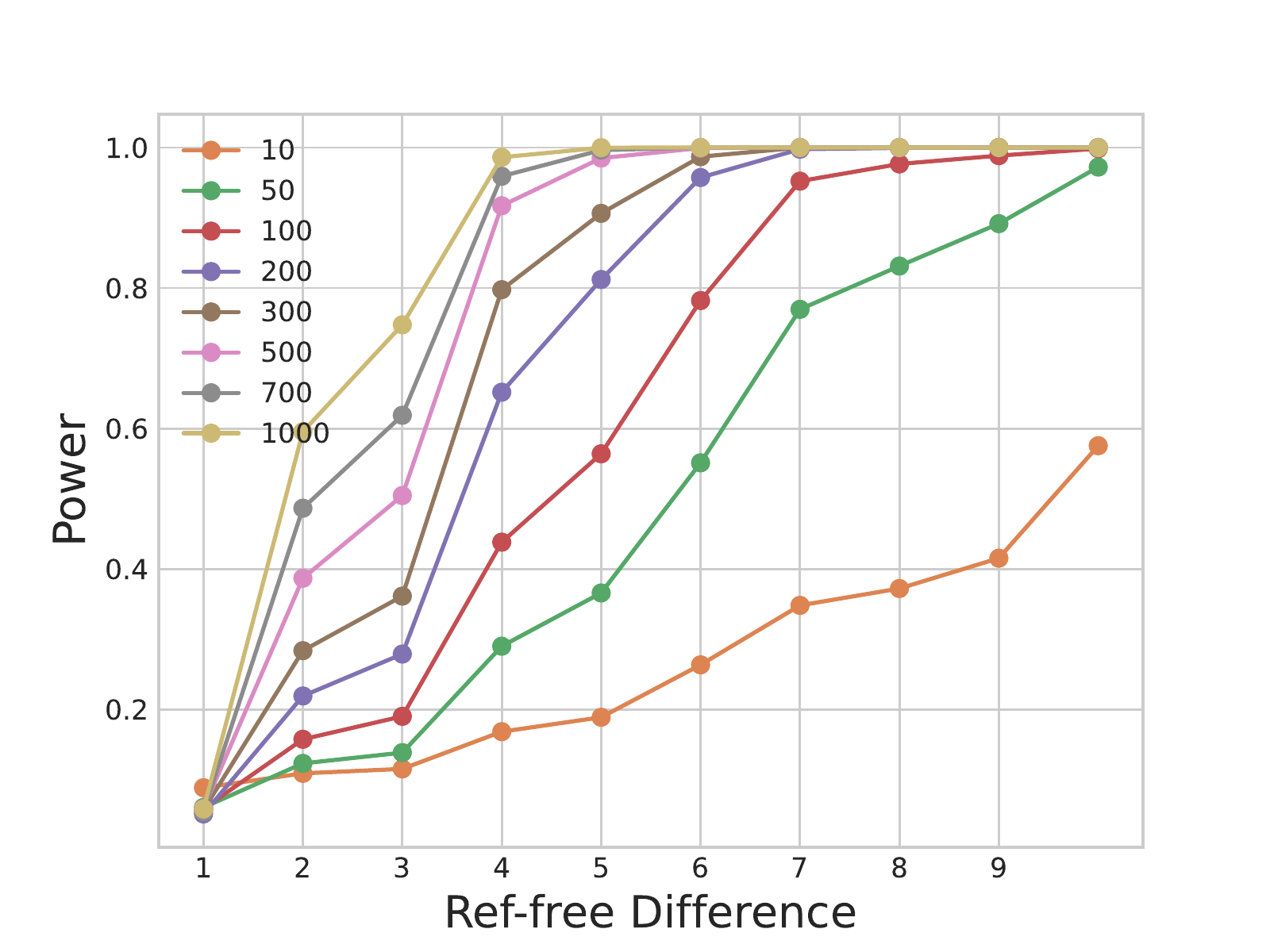}
\caption{\label{subfig:power-reffree}}
    \end{subfigure}
\begin{subfigure}[b]{0.45\textwidth}
         \includegraphics[width=1\linewidth]{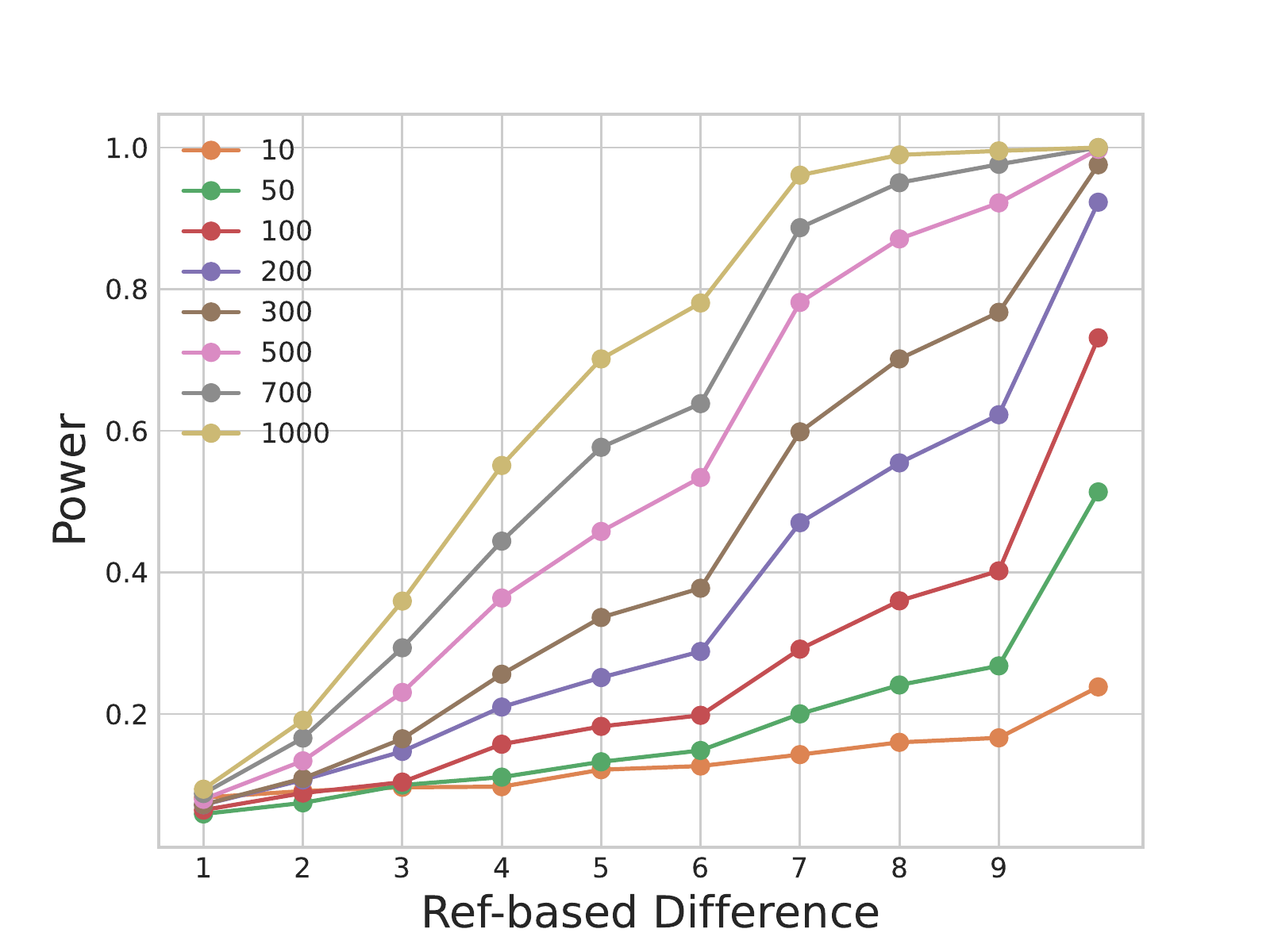}
\caption{\label{subfig:power-refbased}}
    \end{subfigure}
\begin{subfigure}[b]{0.45\textwidth}
         \includegraphics[width=1\linewidth]{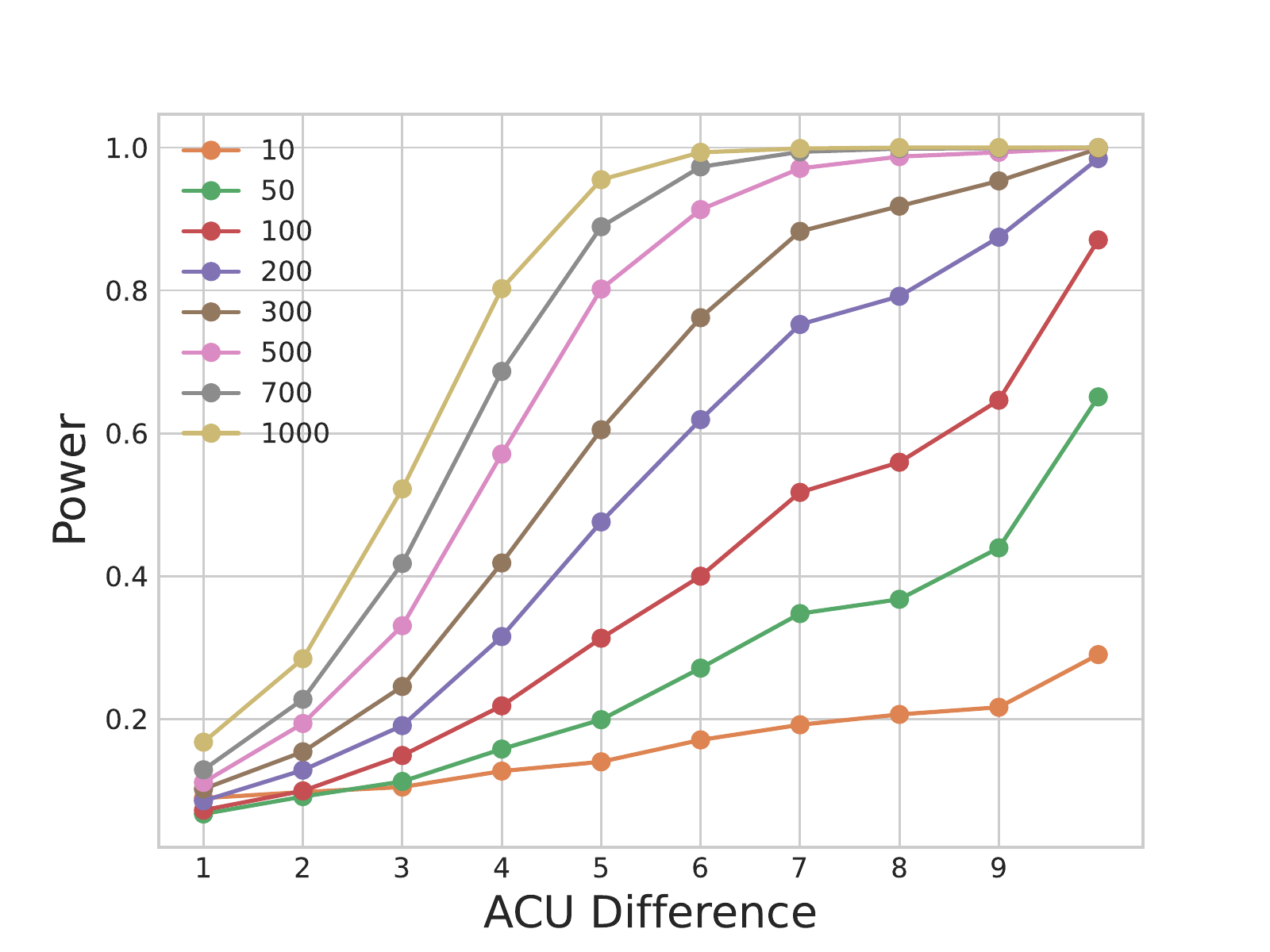}
\caption{\label{subfig:power-acu}}
    \end{subfigure}
 \caption{Power analysis of human evaluation for system comparison under \textit{different evaluation protocols} on the annotated CNNDM test examples. 
 Different lines represent results with different sample sizes.
 The system pairs are grouped into 10 buckets with similar sizes based on their performance difference under human evaluation. 
 Fig.\ref{subfig:power-prior} corresponds to the \textit{\preference} protocol, Fig.\ref{subfig:power-reffree} the \textit{\salience} protocol, Fig.\ref{subfig:power-refbased} the \textit{\accuracy} protocol, and Fig.\ref{subfig:power-acu} the ACU protocol with normalized ACU scores.
 \label{fig:power-protocols}}
\end{figure*}

\paragraph{Summary Level Correlation} 

We show the summary-level Pearson's Correlation Coefficients among different protocols in Tab.~\ref{tab:protocol_corr_summ}.

\paragraph{Power Analysis} 
The power analysis results on the \textit{\preference}, \textit{\salience}, \textit{\accuracy}, and ACU protocols are shown in Fig.~\ref{fig:power-protocols}.

\paragraph{Case Study}

We show a case study in Tab.~\ref{tab:example_annotation_appendix} comparing the summaries generated by BRIO and GPT-3.
GPT-3 scores higher on \textit{\preference} and \textit{\salience} (3.33/3.33 for BRIO and 3.66/4.00 for GPT-3).
However, the BRIO summary scores 0.77 on un-normalized ACU annotations while GPT-3 scores 0.33. 
Also, \textit{\accuracy} annotations favor BRIO over GPT-3 (3.66 vs. 3.33).

\begin{table}[t!]
\small
\centering
\begin{tabular}{lllll}
\toprule
                &  \textbf{\preference}  & \textbf{\salience}  &  \textbf{\accuracy}  & \textbf{\textit{n}ACU}   \\
\midrule
 \preference    & -          & \textbf{0.526}      & 0.056       & 0.082       \\
 \salience      & 0.526      & -          & \textbf{0.070}       & 0.075           \\
 \accuracy      & 0.056      & 0.070      & -           & 0.355   \\
 \textit{n}ACU & 0.082      & 0.075      & 0.355       & -                  \\
\midrule
 Len.        & 0.431      & 0.545      & -0.107      & -0.007              \\
\bottomrule
\end{tabular}
\caption{\textit{Summary-level} Pearson correlations between different protocols on the fine-tuned models. \textbf{\textit{n}ACU} is the normalized ACU score. \textbf{Len.} is the Summary length.
}
\label{tab:protocol_corr_summ} 
\end{table}

\begin{table*}[ht]%h!]
    \centering
    \small
    {
    \begin{tabular}{c}
        \toprule
        \begin{minipage}[t]{0.98\linewidth}
        \textbf{(a) Reference Summary}: 
      Chelsea weren't awarded a penalty for David Ospina's clash with Oscar. Arsenal goalkeeper clattered Oscar inside the box. Brazilian was taken off at half-time, with Didier Drogba replacing him.
 \\
        \end{minipage}\tabularnewline
        \begin{minipage}[t]{0.98\linewidth}
        \textbf{(b) System Summary (BRIO, \citep{liu-etal-2022-brio}):} 
        Oscar collided with Arsenal goalkeeper David Ospina in the 16th minute of the London derby . The Brazilian was substituted at half-time and Jose Mourinho said he suffered `possible concussion'  . Oscar was knocked back by the goalkeeper but Michael Oliver didn't award Chelsea a penalty . \\
        \end{minipage}\tabularnewline 

        \begin{minipage}[t]{0.98\linewidth}
        \textbf{(c) System Summary (GPT-3, \citep{NEURIPS2020_1457c0d6}):} 
        Oscar was forced to leave the match against Arsenal after sustaining a possible concussion from a collision with the opposing goalkeeper.  The referee did not award Chelsea a penalty, despite the collision appearing to warrant one.  Sky Sports pundits agreed that the collision should have been penalized, with some suggesting it could have even warranted a red card. \\
        \end{minipage}\tabularnewline 
        \begin{minipage}[t]{0.98\linewidth}
        \textbf{(d) ACUs with corresponding evaluations}:  
        \vspace{-2\topsep}
        \begin{multicols}{2}
        \begin{itemize*}
      \item Chelsea weren't awarded a penalty.~\textcolor{blue}{\cmark}~\textcolor{green}{\cmark}
\item David Ospina clashed with Oscar.~\textcolor{blue}{\cmark}~\textcolor{green}{\cmark}
\item David Ospina clattered Oscar.~\textcolor{blue}{\cmark}~\textcolor{green}{\cmark}
\item David Ospina plays for Arsenal.~\textcolor{blue}{\cmark}~\textcolor{green}{\xmark}
\item David Ospina is a goalkeeper.~\textcolor{blue}{\cmark}~\textcolor{green}{\xmark}
\item The clash occurred inside the box.~\textcolor{blue}{\xmark}~\textcolor{green}{\xmark}
\item Oscar is Brazilian.~\textcolor{blue}{\cmark}~\textcolor{green}{\xmark}
\item Oscar was taken off at half time.~\textcolor{blue}{\cmark}~\textcolor{green}{\xmark}
\item Didier Drogba replaced Oscar.~\textcolor{blue}{\xmark}~\textcolor{green}{\xmark}
    \end{itemize*}  
    \end{multicols}
    \end{minipage}\tabularnewline \\
    \bottomrule
    \end{tabular}
    }
    \caption{\label{tab:example_annotation_appendix}Example of a reference summary, system summaries
    % from BRIO and GPT-3, 
    and corresponding ACU annotations on CNNDM. The presence or absence of the ACUs for BRIO (in \textcolor{blue}{blue}) and GPT-3 (in \textcolor{green}{green}) are marked by (\cmark) and (\xmark).
    }
% \vspace{-3mm}
\end{table*}

\section{Metric Analysis}
\label{appendix:metric_analysis}

\label{append:metric_analysis}

\subsection{Metrics}
\label{appendix:subsec-metrics}
We provide additional metric details as well as results for other metrics in \S\ref{sec:metric-comparison}. 
Note that for ROUGE, we use the Python implementation.~\footnote{https://pypi.org/project/ROUGE-score/} 
\par 
\noindent \textbf{BLEU}~\citep{papineni-etal-2002-bleu} is a corpus-level precision-focused metric that calculates n-gram overlap and includes a brevity penalty. 
\par
\noindent \textbf{CIDEr}~\citep{vedantam2015cider} computes \{1-4\}-gram co-occurrences, down-weighting common n-grams and calculating cosine similarity between the n-grams of the candidate and reference texts.
\par
\noindent \textbf{Statistics}~\citep{grusky-etal-2018-newsroom} reports summary statistics such as the length, novel and repeated n-grams in the summary, the compression ratio between the summary and article, and measures of the level of extraction. Coverage is the percentage of words that are part of an extractive fragment and density is the average length of the extractive fragment each summary word belongs to.
\par
\noindent \textbf{MoverScore}~\citep{zhao-etal-2019-moverscore} measures semantic distance with Word Mover's Distance~\cite{kusner2015word} on pooled BERT n-gram embeddings.
\par
\noindent \textbf{SUPERT}~\citep{gao-etal-2020-SUPERT} measures the semantic similarity of summaries with pseudo-reference summaries created by extracting salient sentences from the source documents.
\par
\noindent \textbf{BLANC}~\citep{vasilyev-etal-2020-fill}  measures the performance gains of
a pre-trained language model on language understanding tasks on the input document when given access to a document summary.
\par
\noindent \textbf{QAEval}~\citep{deutsch-etal-2021-towards} reports both an F1 and exact match (em) score. 
We do not report the learned answer overlap metric.

\noindent \textbf{SummaQA}~\citep{scialom-etal-2019-answers} reports an F1 score and model confidence. We plan to report QuestEval \cite{scialom-etal-2021-questeval} in a future version.
\par
\noindent \textbf{Lite$^3$Pyramid} includes four variations of the metric depending on the entailment model (two vs three-class entailment model) and how the output is used (as a probability vs a 0/1 label).
\par
\noindent \textbf{CTC}~\citep{deng-etal-2021-Compression} proposes metrics for Compression, transduction, and creation tasks as variations of textual alignment. 
Relevance is scored as the average bi-directional alignment between generated and reference summaries. 
\par
\noindent \textbf{SimCSE}~\citep{gao-etal-2021-SimCSE} apply contrastive learning to learn improved sentence representations, which can then be used to compare generated and reference summary similarity.
\par
\noindent \textbf{UniEval}~\citep{UniEval} frames text evaluation as the answer to yes or no questions, in our case whether the summary is relevant or not, and constructs pseudo-data to fine-tune language models for this setting.

\subsection{Metrics based on Large Language Models}
\label{appendix:llm-metric}
In \S\ref{subsec:metric-eval} we evaluate two different LLM-based automatic evaluation methods.

\textbf{GPTScore}~\citep{Fu2023GPTScoreEA} formulates the text evaluation as the text-filling task and takes the token probability predicted by the LLMs as the quality score.
We use the following prompt for calculating the recall score of the system outputs:
\begin{quote}
    Answer the question based on the following reference summary and candidate summary.
    
Question: Can all of the information in the reference summary be found in the candidate summary? (a). Yes. (b). No.

Reference Summary: \{\{Reference\}\}

Candidate Summary: \{\{Candidate\}\}

Answer: Yes
\end{quote}

The LLM-predicted probability of the last token, ``Yes'', is used as the recall score. 
We use the OpenAI's \texttt{text-davinci-003} as the LLM.

\textbf{G-Eval}~\cite{Liu2023GEvalNE} introduces a similar task as \textbf{GPTscore}, but has the LLM to predict a numerical score directly instead of using the LLM-predicted probability. 
We use the following prompt for the task:
\begin{quote}
    You will receive a reference summary and a candidate summary. Your task is to compare these two summaries and assess the extent to which the candidate summary covers the information presented in the reference summary.

Please indicate your agreement with the following statement:
``All of the information in the reference summary can be found in the candidate summary.''

Use the following 5-point scale when determining your response:

1. Strongly Disagree

2. Disagree

3. Neither Agree nor Disagree

4. Agree

5. Strongly Agree

Input:

Reference Summary:

\{\{Reference\}\}

Candidate Summary:

\{\{Candidate\}\}

Evaluation Form (scores ONLY):

- Agreement (1-5):
\end{quote}

We note that we set the sampling temperature to 0 to ensure more deterministic behavior for G-Eval-3.5 and G-Eval-4.
We also experiment with a sampling strategy with GPT-3.5 (G-Eval-3.5-S), where we sample 5 outputs with a temperate 1 and take the average score as the final prediction.

\subsection{Metric Correlation with ACU Scores}
\label{subsec:appendix-metric-scores}
We collect in total 50 different automatic metrics (including different variations of the same metric), and evaluate their performance using our collected ACU benchmark on CNNDM, XSum and SamSum datasets with three different correlation coefficients (\S\ref{subsec:metric-eval}).
Tab.~\ref{tab:metric-corr-sys-appendix} reports the \textbf{\textit{system-level}} correlation with the \textit{un-normalized} ACU score (Eq.~\ref{eq:acu}).
Tab.~\ref{tab:metric-corr-summ-appendix} reports the \textbf{\textit{summary-level}} correlation with the \textit{un-normalized} ACU score (Eq.~\ref{eq:acu}).
Tab.~\ref{tab:metric-corr-sys-normalized-appendix} reports the \textbf{\textit{system-level}} correlation with the \textit{normalized} ACU score (Eq.~\ref{eq:normalzied-acu}).
Tab.~\ref{tab:metric-corr-summ-normalized-appendix} reports the \textbf{\textit{summary-level}} correlation with the \textit{normalized} ACU score (Eq.~\ref{eq:normalzied-acu}).

\subsection{System Pairs for Fine-grained Metric Evaluation}
\label{subse:system-pairs}

For metric elevation in \S\ref{subsec:metric-eval}, we provide the system pairs in the six different buckets grouped by their performance differences below.

\noindent \textbf{Bucket 1}: CLIFF V.S. FROST, CTRLSUM V.S. GSUM, BART V.S. CLIFF, GOLD V.S. FROST, BART V.S. FROST, CLIFF V.S. GOLD, BRIO V.S. GSUM, GOLD V.S. PEGASUS, BRIO V.S. CTRLSUM, BART V.S. GOLD, BRIO-EXT V.S. MATCHSUM.

\noindent \textbf{Bucket 2}: FROST V.S. PEGASUS, CLIFF V.S. PEGASUS, PEGASUS V.S. GLOB, BRIO-EXT V.S. SIMCLS, BART V.S. PEGASUS, BRIO V.S. MATCHSUM, BART V.S. SIMCLS, GOLD V.S. GLOB, CLIFF V.S. SIMCLS, MATCHSUM V.S. GSUM, SIMCLS V.S. FROST.

\noindent \textbf{Bucket 3}: MATCHSUM V.S. SIMCLS, FROST V.S. GLOB, MATCHSUM V.S. CTRLSUM, CLIFF V.S. GLOB, BRIO V.S. BRIO-EXT, GOLD V.S. SIMCLS, BART V.S. GLOB, BRIO-EXT V.S. GSUM, BRIO-EXT V.S. CTRLSUM, BART V.S. BRIO-EXT, SIMCLS V.S. PEGASUS.

\noindent \textbf{Bucket 4}: CLIFF V.S. BRIO-EXT, BRIO-EXT V.S. FROST, BRIO V.S. SIMCLS, GOLD V.S. BRIO-EXT, BART V.S. MATCHSUM, CLIFF V.S. MATCHSUM, SIMCLS V.S. GSUM, MATCHSUM V.S. FROST, SIMCLS V.S. GLOB, SIMCLS V.S. CTRLSUM, BRIO-EXT V.S. PEGASUS.

\noindent \textbf{Bucket 5}: GOLD V.S. MATCHSUM, MATCHSUM V.S. PEGASUS, BART V.S. BRIO, BRIO-EXT V.S. GLOB, BRIO V.S. CLIFF, BRIO V.S. FROST, BART V.S. GSUM, BART V.S. CTRLSUM, BRIO V.S. GOLD, CLIFF V.S. GSUM, FROST V.S. GSUM.

\noindent \textbf{Bucket 6}: CLIFF V.S. CTRLSUM, MATCHSUM V.S. GLOB, CTRLSUM V.S. FROST, GOLD V.S. GSUM, BRIO V.S. PEGASUS, GOLD V.S. CTRLSUM, PEGASUS V.S. GSUM, CTRLSUM V.S. PEGASUS, BRIO V.S. GLOB, GLOB V.S. GSUM, CTRLSUM V.S. GLOB.

\subsection{Confidence Interval}
\label{subsec:ci-appendix}
\begin{figure}[t!]
\centering
\includegraphics[width=0.9\linewidth]{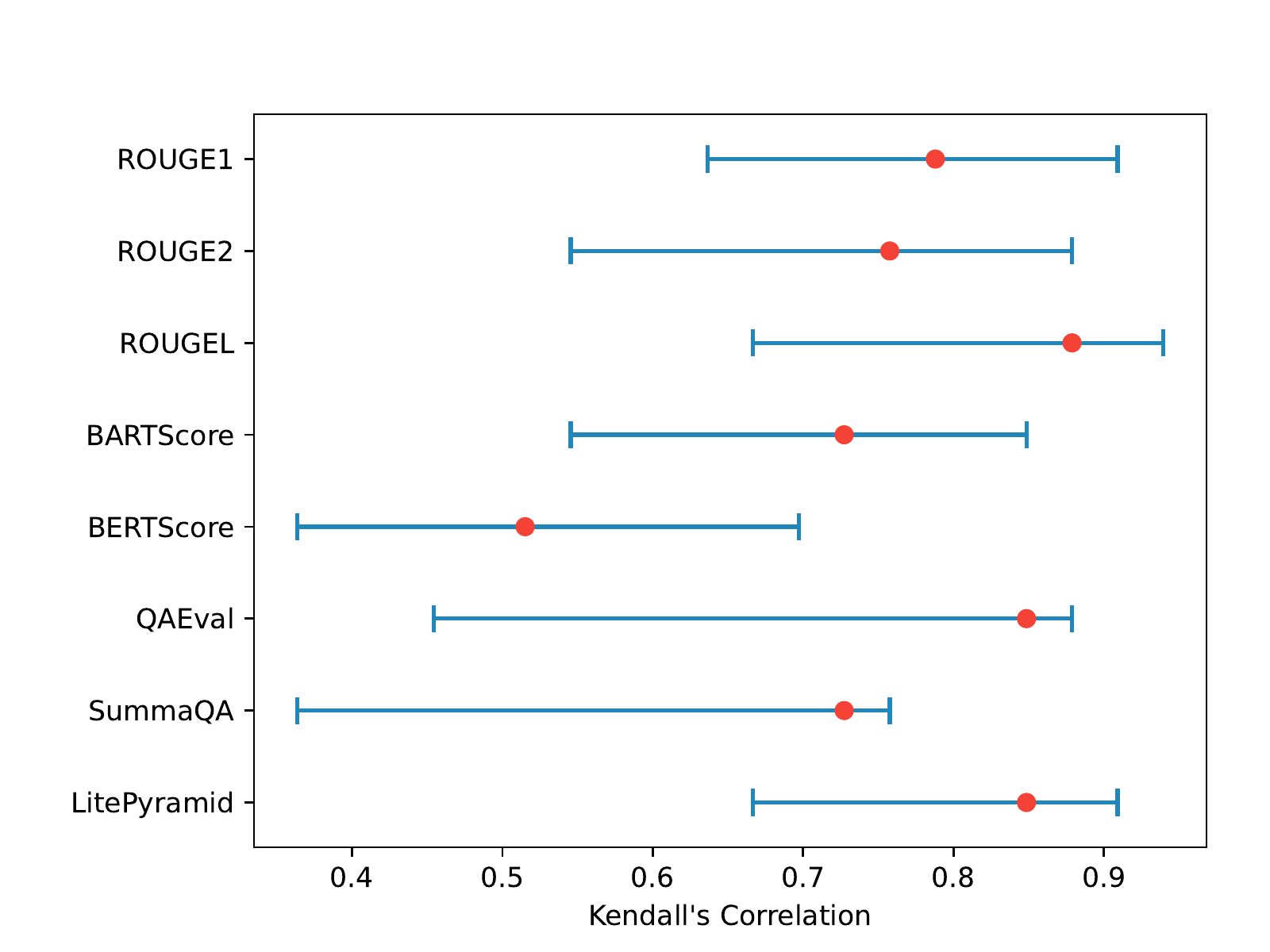}
 \caption{Confidence intervals of the system-level Kendall's correlation coefficients between automatic metrics and ACU scores.}
 \label{fig:ci}
\end{figure}

\begin{figure}[t!]
\centering
\includegraphics[width=1\linewidth]{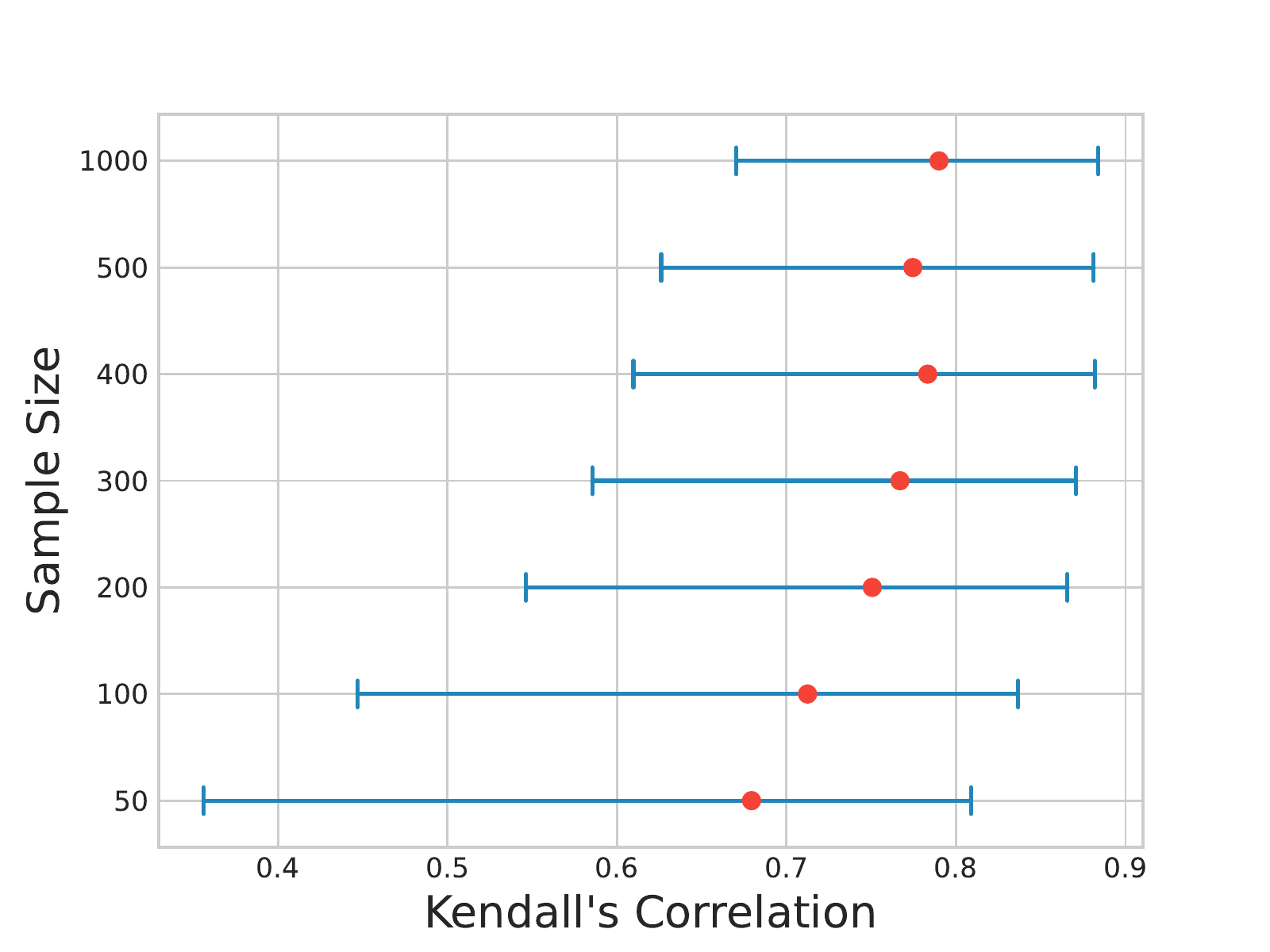}
 \caption{Confidence intervals of the system-level Kendall's correlation coefficients between ROUGE1 recall scores and ACU scores under different sample sizes.}
 \label{fig:ci_sample}
\end{figure}

We select several automatic metrics and calculate the confidence intervals of their system-level correlations with the ACU scores (\S\ref{subsec:metric-analysis}). 
The results are in Fig.~\ref{fig:ci}.
Similar to \citet{deutsch-etal-2021-statistical}, we found that the confidence intervals are large. 
However, having a larger sample size can effectively reduce the confidence interval.
Specifically, we use re-sampling to generate a series of synthetic sample sets with several different sizes and calculate the confidence interval by averaging over the sampled sets with the same size. 
As shown in Fig.~\ref{fig:ci_sample}, larger sample sizes lead to more stable results.

\subsection{Power Analysis of Metric Comparison}
\label{subsec:power-analysis-metric-appendix}
We use Alg.\ref{alg:power-analysis} to conduct a power analysis of metric comparison based on their Kendall's correlations with ACU scores (\S\ref{subsec:metric-analysis}).
We choose 20 metrics for comparison, resulting in 190 metric pairs in total, which are (1) BARTScore-r-parabank, (2) BERTScore-r-deberta, (3) BERTScore-r-roberta, (4) BLANC, (5) CHRF, (6) CTC, (7) Meteor, (8) Lite$^2$Pyramid-p2c, (9) QAEval-em, (10) QAEval-f1, (11) ROUGE1, (12) ROUGE1r, (13) ROUGE2, (14) ROUGE2r, (15) ROUGEL, (16) ROUGELr, (17) SimCSE, (18) SummaQA, (19) SummaQA-prob, (20) SUPERT.
We note that we use the \textit{permutation test} instead of the \textit{paired bootstrapping test} to calculate the statistical significance for metric comparison, since \citet{deutsch-etal-2021-statistical} found that the permutation test works better for detecting significant results in metric comparison.

\subsection{Metric Correlation with Different Human Evaluation Protocols}
\label{subsec:corr-protocol-appendix}

We present the correlations between automatic metrics and different human evaluation protocols in Tab.~\ref{tab:metric-corr-protocol-appendix} as discussed in \S\ref{subsec:metric-analysis}.

\begin{table*}[t!]
\small
\centering

\begin{tabular}{lcccc}
\toprule
 \textbf{Best Practice \& Implementation} &   \textbf{Yes} &   \textbf{No} &   \textbf{Percentage (\%)} \\
\midrule
\textbf{Performed Human Evaluation}                   & 39     & 16     & 71    \\
\midrule
\textbf{Produce Robust Human Evaluation Result}    &  &  &\textbf{(out of 39)}  \\
Performed Significance Test                  & 27     & 12     & 69  \\
Performed Power Analysis                  & 0     & 39     & 0    \\
Reported Inter-annotator Agreement                   & 12     & 27     & 28    \\
~~~Document Specific Agreement Test                  & 9     & 3     & 75   \\ 
~~~Document Agreement Value                      & 12    &0      &100 \\
\midrule
\textbf{Documentation of the study setup (questionnaire, sample answers, platform, etc.)  }              & 39     & 0     & 100    \\
Participants  (crowd-worker, expert, etc.)                 & 37     & 2     & 97    \\
Document Sample Size                      & 29    &10      &74\\
Document Participant Number                      & 35    &4      &90\\

Document Participant Demographics                      & 24    &15      &62\\
\midrule
\textbf{Released Human Evaluation Data}
                 & 4     & 35     & 10    \\
\bottomrule
\end{tabular}
\caption{Survey of human evaluation practices in recent text summarization research.}
\label{tab:survey} 
\end{table*}

\section{Human Evaluation Practices in Recent Text Summarization Research}
\label{appendix:survey}

\begin{figure}[t!]
\centering
\includegraphics[width=0.9\linewidth]{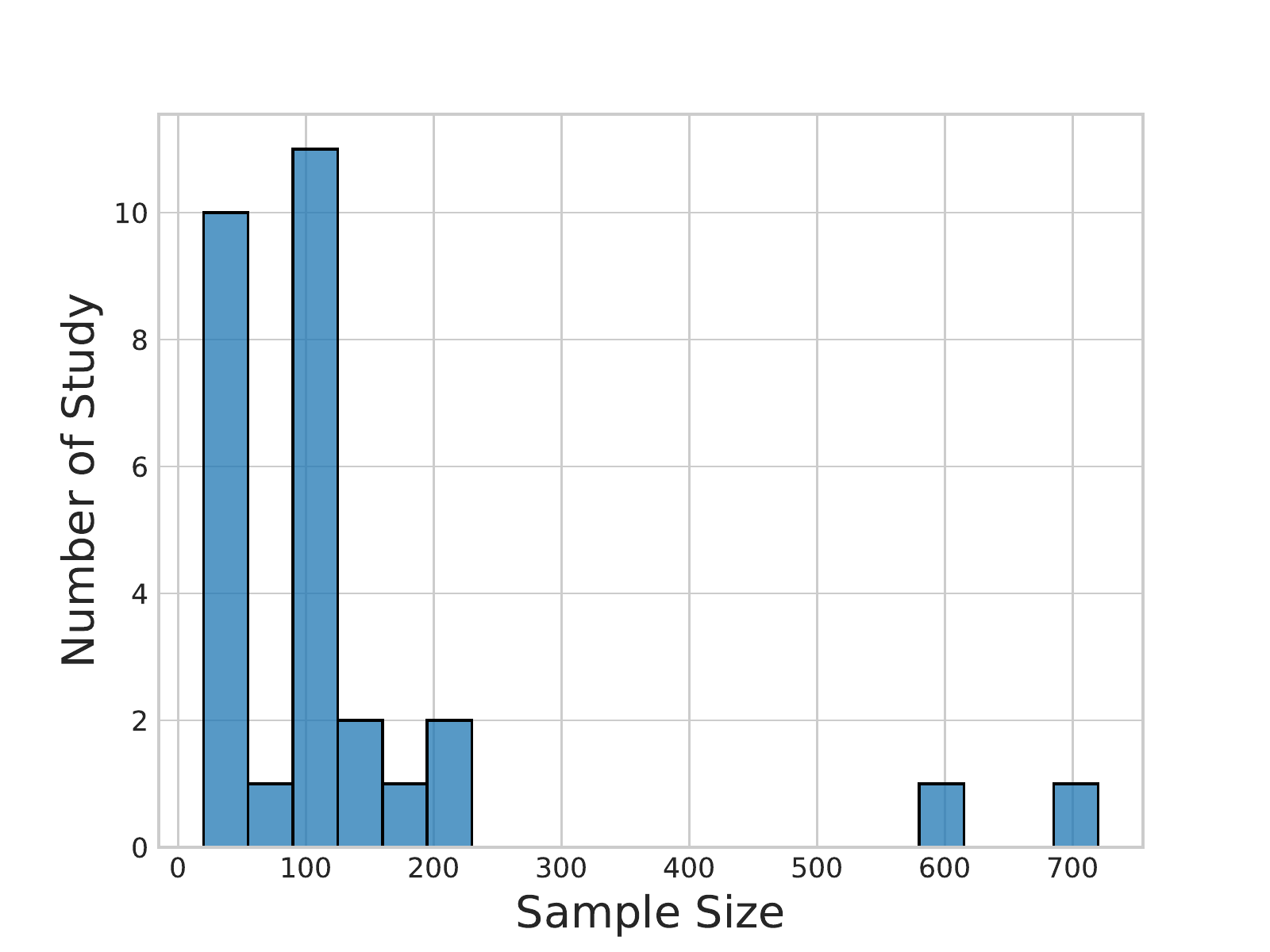}
 \caption{Sample size of the conducted human evaluation study in recent text summarization research based on our survey (Appendix~\ref{appendix:survey}).}
 \label{fig:survey}
\end{figure}

We provide a brief survey for the human evaluation practices of 55 selected papers on text summarization published at NAACL\footnote{\url{https://aclanthology.org/events/naacl-2022/}}, ACL\footnote{\url{https://aclanthology.org/events/acl-2022/}}, and EMNLP\footnote{\url{https://preview.aclanthology.org/emnlp-22-ingestion/volumes/2022.emnlp-main/}} from 2022. 
We follow the design of a similar study in \citet{Gehrmann2022RepairingTC} as described below.
The results are shown in Tab. \ref{tab:survey}.

\noindent \textbf{Performed Human Evaluation:} Report ``yes'', if a human evaluation of any kind is done. We report that 71\% of analyzed papers did human evaluation.

\noindent \textbf{Significance Test:} Report ``yes'', if a significance test is done on the human annotation results. 
Of the 39 papers that conducted human evaluation, a total of 27 papers reported the result of a significance test (68\%), which is much higher compared to the 25\% reported in the previous survey \cite{Gehrmann2022RepairingTC}.

\noindent \textbf{Power Analysis:} Report ``yes'', if a power analysis of any kind is mentioned. 
Of the 39 papers that conducted human evaluation, none of the papers did power analysis, the same as the result provided in the previous survey of \citet{Gehrmann2022RepairingTC}.
\noindent \textbf{Inter-annotator Agreement:} Report ``yes'', if any kind of agreement test is conducted to evaluate the quality of human annotation themselves. Overall, we report a total of 12 papers (28\%) that did agreement tests and documented specific agreement values. 9 out of the 12 papers recorded the specific agreement test, with Krippendorff’s alpha as the most commonly used measurement. 

\noindent \textbf{Participants (crowd-worker, expert, etc.)} Report ``yes'', if at least the number of human evaluators, document sample size, annotators per document, or their demographics is mentioned.
We show the sample size of the conducted human evaluation study in Fig.~\ref{fig:survey}, and note that around 93\% of them are less or equal to 200.

\noindent \textbf{Released Human Evaluation Data:} Report ``yes'', if the authors release the human evaluation data.

\begin{table*}[t!]
\small
\centering
\addtolength{\tabcolsep}{-1pt} 
\begin{tabular}{@{\extracolsep{1pt}}lccccccccc@{}}
\toprule
 & \multicolumn{3}{c}{\textbf{CNNDM}} & \multicolumn{3}{c}{\textbf{XSum}} & \multicolumn{3}{c}{\textbf{SamSum}} \\
  & $r$ & $\rho$ & $\tau$ &  $r$ & $\rho$ & $\tau$ &  $r$ & $\rho$ & $\tau$ \\
 \cmidrule{2-4} \cmidrule{5-7} \cmidrule{8-10} 
  BARTScore-f1-cnndm                 & 0.132           & 0.119            & 0.000           & -0.198         & 0.095           & 0.071          & 0.907            & 0.952             & 0.857            \\
 BARTScore-f1-parabank              & 0.219           & 0.070            & 0.000           & 0.428          & 0.429           & 0.429          & 0.881            & 0.976             & 0.929            \\
 BARTScore-p-cnndm          & -0.036          & -0.189           & -0.121          & -0.650         & -0.571          & -0.429         & 0.777            & 0.571             & 0.429            \\
 BARTScore-p-parabank       & -0.187          & -0.294           & -0.212          & -0.387         & -0.452          & -0.286         & 0.692            & 0.524             & 0.286            \\
 BARTScore-r-cnndm             & 0.909           & 0.881            & 0.727           & 0.868          & 0.786           & 0.643          & 0.945            & 0.976             & 0.929            \\
 BARTScore-r-parabank          & 0.891           & 0.902            & 0.727           & 0.920          & \textbf{0.881}           & \textbf{0.714}          & 0.932            & 0.976             & 0.929            \\
 BERTScore-f1-deberta               & 0.062           & 0.119            & 0.000           & 0.543          & 0.429           & 0.429          & 0.849            & 0.786             & 0.643            \\
 BERTScore-f1-roberta               & 0.103           & 0.028            & -0.091          & 0.592          & 0.429           & 0.429          & 0.852            & 0.809             & 0.714            \\
 BERTScore-p-deberta                & -0.439          & -0.510           & -0.394          & 0.129          & 0.405           & 0.357          & 0.373            & 0.405             & 0.214            \\
 BERTScore-p-roberta                & -0.275          & -0.350           & -0.273          & 0.172          & 0.333           & 0.286          & 0.374            & 0.381             & 0.214            \\
 BERTScore-r-deberta                & 0.649           & 0.552            & 0.424           & 0.878          & 0.762           & 0.571          & 0.951            & 0.952             & 0.857            \\
 BERTScore-r-roberta                & 0.750           & 0.713            & 0.515           & 0.920          & 0.786           & 0.571          & 0.939            & 0.952             & 0.857            \\
 BLANC                              & 0.588           & 0.699            & 0.515           & 0.065          & -0.024          & 0.071          & 0.824            & 0.809             & 0.714            \\
 BLEU                               & -0.184          & -0.273           & -0.212          & 0.631          & 0.595           & 0.571          & 0.806            & 0.833             & 0.714            \\
 CHRF                               & 0.894           & 0.916            & 0.758           & 0.883          & 0.762           & 0.571          & 0.937            & 0.952             & 0.857            \\
 Compression                        & -0.711          & -0.769           & -0.606          & -0.185         & 0.071           & 0.000          & -0.699           & -0.762            & -0.571           \\
 Coverage                           & -0.013          & -0.168           & 0.000           & -0.568         & -0.571          & -0.429         & 0.719            & 0.809             & 0.643            \\
 CTC                                & 0.516           & 0.692            & 0.485           & 0.399          & 0.405           & 0.214          & 0.964            & 0.976             & 0.929            \\
 Density                            & 0.201           & 0.161            & 0.151           & -0.415         & -0.381          & -0.286         & 0.815            & 0.762             & 0.571            \\
 Lite$^3$Pyramid-l2c                                & 0.950           & 0.958            & 0.849           & 0.903          & 0.809           & 0.643          & 0.984            & \textbf{1.000}             & \textbf{1.000}            \\
 Lite$^3$Pyramid-l3c                                & 0.952           & 0.951            & 0.849           & 0.914          & 0.809           & 0.643          & \textbf{0.989}            & \textbf{1.000}             & \textbf{1.000}            \\
  Lite$^3$Pyramid-p2c                                & \textbf{0.953}           & 0.958            & 0.849           & 0.914          & 0.833           & \textbf{0.714}          & 0.986            & \textbf{1.000}             & \textbf{1.000}            \\
 Lite$^3$Pyramid-p3c                                & 0.950           & \textbf{0.965}            & \textbf{0.879}           & \textbf{0.927}          & 0.809           & 0.643          & 0.987            & \textbf{1.000}             & \textbf{1.000}            \\
 Meteor                             & 0.909           & 0.916            & 0.758           & 0.905          & 0.762           & 0.571          & 0.911            & 0.952             & 0.857            \\
 MoverScore                         & -0.173          & -0.161           & -0.121          & 0.674          & 0.571           & 0.500          & 0.820            & 0.833             & 0.714            \\
 Novel-1gram            & 0.072           & 0.224            & 0.000           & 0.608          & 0.452           & 0.357          & -0.740           & -0.809            & -0.643           \\
 Novel-2gram            & -0.013          & 0.063            & 0.000           & 0.578          & 0.619           & 0.429          & -0.843           & -0.833            & -0.643           \\
 Repeated-1gram & 0.723           & 0.643            & 0.333           & 0.172          & 0.095           & 0.143          & 0.399            & 0.286             & 0.214            \\
 Repeated-2gram & 0.499           & 0.294            & 0.151           & 0.257          & 0.119           & 0.143          & -0.277           & -0.024            & 0.000            \\
 QAEval-em                          & 0.723           & 0.629            & 0.515           & 0.450          & 0.452           & 0.357          & 0.947            & 0.952             & 0.857            \\
 QAEval-f1                          & 0.925           & 0.944            & 0.849           & 0.551          & 0.500           & 0.429          & 0.962            & 0.976             & 0.929            \\
 ROUGE1                             & 0.382           & 0.301            & 0.151           & 0.665          & 0.476           & 0.357          & 0.942            & \textbf{1.000}             & \textbf{1.000}            \\
 ROUGE1p                            & -0.490          & -0.503           & -0.394          & 0.195          & 0.405           & 0.357          & 0.164            & 0.191             & 0.071            \\
 ROUGE1r                            & 0.947           & 0.937            & 0.788           & 0.767          & 0.857           & \textbf{0.714}          & 0.920            & 0.976             & 0.929            \\
 ROUGE2                             & 0.236           & 0.063            & 0.000           & 0.620          & 0.500           & 0.429          & 0.889            & 0.905             & 0.786            \\
 ROUGE2p                            & -0.412          & -0.455           & -0.333          & 0.323          & 0.429           & 0.429          & 0.535            & 0.571             & 0.357            \\
 ROUGE2r                            & 0.923           & 0.909            & 0.758           & 0.888          & 0.786           & 0.643          & 0.977            & \textbf{1.000}             & \textbf{1.000}            \\
 ROUGEL                             & 0.206           & 0.091            & -0.030          & 0.572          & 0.429           & 0.429          & 0.915            & 0.976             & 0.929            \\
 ROUGELp                            & -0.483          & -0.566           & -0.424          & 0.181          & 0.357           & 0.286          & 0.317            & 0.381             & 0.214            \\
 ROUGELr                            & 0.944           & 0.958            & \textbf{0.879}           & 0.836          & 0.809           & 0.643          & 0.932            & 0.976             & 0.929            \\
 SimCSE                             & 0.816           & 0.853            & 0.636           & 0.865          & 0.809           & 0.571          & 0.924            & \textbf{1.000}             & \textbf{1.000}            \\
 SummaQA                 & 0.810           & 0.853            & 0.697           & -0.199         & -0.119          & 0.000          & 0.717            & 0.595             & 0.429            \\
 SummaQA-prob                   & 0.749           & 0.860            & 0.727           & 0.308          & 0.214           & 0.143          & 0.817            & 0.738             & 0.643            \\
 Summary-length                     & 0.780           & 0.818            & 0.667           & 0.226          & -0.071          & 0.000          & 0.699            & 0.762             & 0.571            \\
 SUPERT                             & 0.406           & 0.552            & 0.424           & 0.004          & -0.095          & -0.071         & 0.673            & 0.691             & 0.429            \\
 UniEval-coherence                  & -0.325          & 0.126            & 0.000           & 0.095          & 0.095           & 0.071          & 0.702            & 0.786             & 0.643            \\
 UniEval-consistency                & 0.001           & -0.168           & -0.061          & 0.056          & 0.071           & 0.071          & 0.344            & 0.238             & 0.071            \\
 UniEval-fluency                    & 0.249           & 0.420            & 0.273           & -0.703         & -0.643          & -0.500         & 0.405            & 0.381             & 0.286            \\
 UniEval-overall                    & -0.201          & 0.028            & 0.030           & -0.062         & 0.048           & 0.071          & 0.739            & 0.571             & 0.500            \\
 UniEval-relevance                  & -0.148          & 0.119            & 0.091           & 0.017          & 0.333           & 0.214          & 0.742            & 0.643             & 0.500            \\
 \bottomrule
\end{tabular}
\addtolength{\tabcolsep}{1pt} 
\caption{The \textbf{\textit{system-level}} Pearson's $r$, Spearman's $\rho$, and Kendall's $\tau$ correlation coefficients between the automatic metric scores and \textbf{\textit{un-normalized}} ACU scores of system outputs on CNNDM, XSum and SamSum datasets.
}
\label{tab:metric-corr-sys-appendix} 
\end{table*}

\begin{table*}[t!]
\small
\centering
\addtolength{\tabcolsep}{-1pt} 
\begin{tabular}{@{\extracolsep{1pt}}lccccccccc@{}}
\toprule
 & \multicolumn{3}{c}{\textbf{CNNDM}} & \multicolumn{3}{c}{\textbf{XSum}} & \multicolumn{3}{c}{\textbf{SamSum}} \\
 % \\
  & $r$ & $\rho$ & $\tau$ &  $r$ & $\rho$ & $\tau$ &  $r$ & $\rho$ & $\tau$ \\
   \cmidrule{2-4} \cmidrule{5-7} \cmidrule{8-10} 
   BARTScore-f1-cnndm                 & 0.353           & 0.329            & 0.264           & 0.261          & 0.238           & 0.202          & 0.434            & 0.401             & 0.340            \\
 BARTScore-f1-parabank              & 0.417           & 0.386            & 0.311           & 0.309          & 0.279           & 0.239          & 0.430            & 0.396             & 0.340            \\
 BARTScore-p-cnndm          & 0.178           & 0.161            & 0.128           & 0.188          & 0.166           & 0.140          & 0.282            & 0.263             & 0.224            \\
 BARTScore-p-parabank       & 0.237           & 0.216            & 0.170           & 0.235          & 0.220           & 0.187          & 0.269            & 0.249             & 0.212            \\
 BARTScore-r-cnndm             & 0.567           & 0.530            & 0.435           & 0.325          & 0.300           & 0.260          & 0.546            & 0.508             & 0.438            \\
 BARTScore-r-parabank          & 0.582           & 0.548            & 0.453           & \textbf{0.353}          & \textbf{0.326}           & 0.282          & 0.531            & 0.500             & 0.430            \\
 BERTScore-f1-deberta               & 0.441           & 0.413            & 0.334           & 0.290          & 0.280           & 0.241          & 0.401            & 0.377             & 0.326            \\
 BERTScore-f1-roberta               & 0.432           & 0.397            & 0.320           & 0.305          & 0.285           & 0.244          & 0.415            & 0.388             & 0.335            \\
 BERTScore-p-deberta                & 0.255           & 0.239            & 0.191           & 0.209          & 0.211           & 0.180          & 0.208            & 0.204             & 0.173            \\
 BERTScore-p-roberta                & 0.218           & 0.200            & 0.160           & 0.223          & 0.221           & 0.190          & 0.212            & 0.209             & 0.178            \\
 BERTScore-r-deberta                & 0.544           & 0.516            & 0.424           & 0.327          & 0.305           & 0.262          & 0.507            & 0.476             & 0.409            \\
 BERTScore-r-roberta                & 0.571           & 0.542            & 0.448           & 0.348          & 0.320           & 0.277          & 0.516            & 0.481             & 0.417            \\
 BLANC                              & 0.238           & 0.220            & 0.175           & -0.018         & -0.022          & -0.020         & 0.167            & 0.156             & 0.136            \\
 BLEU                               & 0.337           & 0.306            & 0.246           & 0.275          & 0.259           & 0.227          & 0.373            & 0.356             & 0.306            \\
 CHRF                               & 0.564           & 0.528            & 0.436           & 0.353          & 0.315           & 0.275          & 0.486            & 0.459             & 0.396            \\
 Compression                        & -0.309          & -0.296           & -0.238          & -0.088         & -0.080          & -0.071         & -0.312           & -0.307            & -0.269           \\
 Coverage                           & 0.012           & 0.005            & 0.003           & -0.045         & -0.044          & -0.037         & 0.056            & 0.044             & 0.037            \\
 CTC                                & 0.453           & 0.431            & 0.348           & 0.270          & 0.249           & 0.215          & 0.476            & 0.442             & 0.382            \\
 Density                            & 0.078           & 0.070            & 0.054           & -0.054         & -0.052          & -0.044         & 0.119            & 0.109             & 0.091            \\
 Lite$^3$Pyramid-l2c                                & 0.537           & 0.523            & 0.466           & 0.219          & 0.219           & 0.207          & 0.524            & 0.519             & 0.494            \\
 Lite$^3$Pyramid-l3c                                & 0.532           & 0.521            & 0.466           & 0.217          & 0.214           & 0.204          & 0.540            & 0.535             & \textbf{0.509}            \\
  Lite$^3$Pyramid-p2c                                & 0.582           & 0.546            & 0.452           & 0.303          & 0.284           & 0.245          & 0.599            & 0.539             & 0.467            \\
 Lite$^3$Pyramid-p3c                                & \textbf{0.584}           & 0.543            & 0.448           & 0.310          & 0.285           & 0.246          & \textbf{0.615}            & \textbf{0.549}             & 0.475            \\
 Meteor                             & 0.537           & 0.496            & 0.407           & 0.327          & 0.308           & 0.268          & 0.471            & 0.430             & 0.373            \\
 MoverScore                         & 0.388           & 0.364            & 0.292           & 0.320          & 0.296           & 0.252          & 0.398            & 0.375             & 0.320            \\
 Novel-1gram            & -0.008          & -0.005           & -0.003          & 0.051          & 0.048           & 0.041          & -0.070           & -0.066            & -0.056           \\
 Novel-2gram            & -0.026          & -0.035           & -0.028          & 0.057          & 0.057           & 0.050          & -0.112           & -0.103            & -0.087           \\
 Repeated-1gram & 0.071           & 0.067            & 0.052           & 0.010          & 0.006           & 0.005          & 0.172            & 0.172             & 0.152            \\
 Repeated-2gram & 0.061           & 0.060            & 0.048           & 0.010          & 0.006           & 0.005          & 0.059            & 0.057             & 0.052            \\
 QAEval-em                          & 0.350           & 0.334            & 0.296           & 0.159          & 0.156           & 0.149          & 0.383            & 0.377             & 0.352            \\
 QAEval-f1                          & 0.454           & 0.427            & 0.358           & 0.226          & 0.215           & 0.198          & 0.437            & 0.421             & 0.384            \\
 ROUGE1                             & 0.457           & 0.430            & 0.348           & 0.302          & 0.292           & 0.253          & 0.416            & 0.398             & 0.345            \\
 ROUGE1p                            & 0.190           & 0.175            & 0.140           & 0.227          & 0.224           & 0.194          & 0.113            & 0.119             & 0.103            \\
 ROUGE1r                            & 0.579           & \textbf{0.552}            & \textbf{0.468}           & 0.328          & 0.322           & \textbf{0.293}          & 0.503            & 0.485             & 0.439            \\
 ROUGE2                             & 0.444           & 0.407            & 0.329           & 0.277          & 0.255           & 0.222          & 0.380            & 0.350             & 0.301            \\
 ROUGE2p                            & 0.307           & 0.287            & 0.229           & 0.241          & 0.229           & 0.200          & 0.214            & 0.210             & 0.181            \\
 ROUGE2r                            & 0.552           & 0.529            & 0.453           & 0.301          & 0.291           & 0.266          & 0.456            & 0.436             & 0.395            \\
 ROUGEL                             & 0.430           & 0.399            & 0.321           & 0.266          & 0.249           & 0.215          & 0.395            & 0.372             & 0.323            \\
 ROUGELp                            & 0.192           & 0.179            & 0.143           & 0.214          & 0.208           & 0.180          & 0.121            & 0.120             & 0.103            \\
 ROUGELr                            & 0.561           & 0.537            & 0.454           & 0.297          & 0.285           & 0.258          & 0.480            & 0.460             & 0.415            \\
 SimCSE                             & 0.461           & 0.429            & 0.346           & 0.308          & 0.290           & 0.248          & 0.450            & 0.420             & 0.360            \\
 SummaQA                 & 0.165           & 0.153            & 0.121           & 0.022          & 0.015           & 0.013          & 0.045            & 0.049             & 0.039            \\
 SummaQA-prob                   & 0.155           & 0.150            & 0.119           & 0.026          & 0.023           & 0.019          & 0.131            & 0.120             & 0.102            \\
 Summary-length                     & 0.315           & 0.296            & 0.238           & 0.081          & 0.075           & 0.067          & 0.314            & 0.307             & 0.268            \\
 SUPERT                             & 0.211           & 0.206            & 0.165           & 0.047          & 0.049           & 0.042          & 0.191            & 0.168             & 0.141            \\
 UniEval-coherence                  & 0.098           & 0.127            & 0.100           & 0.017          & 0.011           & 0.012          & 0.186            & 0.197             & 0.167            \\
 UniEval-consistency                & 0.007           & 0.015            & 0.010           & 0.017          & 0.013           & 0.013          & 0.044            & 0.037             & 0.031            \\
 UniEval-fluency                    & -0.008          & -0.022           & -0.015          & -0.031         & -0.040          & -0.034         & -0.006           & -0.035            & -0.028           \\
 UniEval-overall                    & 0.111           & 0.132            & 0.104           & 0.089          & 0.078           & 0.067          & 0.171            & 0.187             & 0.157            \\
 UniEval-relevance                  & 0.129           & 0.154            & 0.121           & 0.180          & 0.181           & 0.152          & 0.201            & 0.235             & 0.197            \\
 \bottomrule
\end{tabular}
\addtolength{\tabcolsep}{1pt} 
\caption{The \textbf{\textit{summary-level}} Pearson's $r$, Spearman's $\rho$, and Kendall's $\tau$ correlation coefficients between the automatic metric scores and \textbf{\textit{un-normalized}} ACU scores of system outputs on CNNDM, XSum and SamSum.
}
\label{tab:metric-corr-summ-appendix} 
\end{table*}

\begin{table*}[t!]
\small
\centering
\addtolength{\tabcolsep}{-1pt} 
\begin{tabular}{@{\extracolsep{1pt}}lccccccccc@{}}
\toprule
 & \multicolumn{3}{c}{\textbf{CNNDM}} & \multicolumn{3}{c}{\textbf{XSum}} & \multicolumn{3}{c}{\textbf{SamSum}} \\
 %  \\
  & $r$ & $\rho$ & $\tau$ &  $r$ & $\rho$ & $\tau$ &  $r$ & $\rho$ & $\tau$ \\
 \cmidrule{2-4} \cmidrule{5-7} \cmidrule{8-10} 
  BARTScore-f1-cnndm                 & 0.539           & 0.706            & 0.455           & -0.148         & 0.095           & 0.071          & 0.920            & 0.786             & 0.643            \\
 BARTScore-f1-parabank              & 0.692           & 0.706            & 0.455           & 0.478          & 0.429           & 0.429          & 0.907            & 0.714             & 0.571            \\
 BARTScore-p-cnndm          & 0.430           & 0.524            & 0.333           & -0.622         & -0.571          & -0.429         & 0.945            & 0.738             & 0.643            \\
 BARTScore-p-parabank       & 0.421           & 0.413            & 0.242           & -0.341         & -0.452          & -0.286         & 0.915            & 0.691             & 0.500            \\
 BARTScore-r-cnndm             & 0.461           & 0.364            & 0.273           & 0.893          & 0.786           & 0.643          & 0.774            & 0.738             & 0.571            \\
 BARTScore-r-parabank          & 0.756           & 0.615            & 0.455           & \textbf{0.932}          & \textbf{0.881}           & \textbf{0.714}          & 0.786            & 0.738             & 0.571            \\
 BERTScore-f1-deberta               & 0.643           & 0.783            & 0.576           & 0.593          & 0.429           & 0.429          & \textbf{0.985}            & 0.952             & 0.857            \\
 BERTScore-f1-roberta               & 0.733           & 0.755            & 0.545           & 0.639          & 0.429           & 0.429          & 0.955            & \textbf{0.976}             & \textbf{0.929}            \\
 BERTScore-p-deberta                & 0.335           & 0.378            & 0.242           & 0.191          & 0.405           & 0.357          & 0.756            & 0.619             & 0.571            \\
 BERTScore-p-roberta                & 0.432           & 0.420            & 0.242           & 0.233          & 0.333           & 0.286          & 0.779            & 0.667             & 0.571            \\
 BERTScore-r-deberta                & 0.664           & 0.489            & 0.333           & 0.863          & 0.762           & 0.571          & 0.845            & 0.714             & 0.500            \\
 BERTScore-r-roberta                & 0.725           & 0.552            & 0.364           & 0.909          & 0.786           & 0.571          & 0.802            & 0.714             & 0.500            \\
 BLANC                              & -0.122          & -0.126           & -0.061          & 0.020          & -0.024          & 0.071          & 0.506            & 0.452             & 0.357            \\
 BLEU                               & 0.442           & 0.482            & 0.303           & 0.676          & 0.595           & 0.571          & 0.906            & 0.905             & 0.786            \\
 CHRF                               & 0.701           & 0.601            & 0.424           & 0.869          & 0.762           & 0.571          & 0.818            & 0.714             & 0.500            \\
 Compression                        & -0.099          & -0.077           & -0.091          & -0.128         & 0.071           & 0.000          & -0.361           & -0.357            & -0.214           \\
 Coverage                           & -0.599          & -0.797           & -0.576          & -0.603         & -0.571          & -0.429         & 0.606            & 0.381             & 0.286            \\
 CTC                                & -0.074          & -0.035           & -0.030          & 0.350          & 0.405           & 0.214          & 0.792            & 0.738             & 0.571            \\
 Density                            & -0.366          & -0.685           & -0.485          & -0.427         & -0.381          & -0.286         & 0.574            & 0.286             & 0.214            \\
 Lite$^3$Pyramid-l2c                                & 0.501           & 0.462            & 0.273           & 0.903          & 0.809           & 0.643          & 0.856            & 0.786             & 0.643            \\
 Lite$^3$Pyramid-l3c                                & 0.486           & 0.448            & 0.273           & 0.908          & 0.809           & 0.643          & 0.845            & 0.786             & 0.643            \\
  Lite$^3$Pyramid-p2c                                & 0.510           & 0.462            & 0.273           & 0.915          & 0.833           & \textbf{0.714}          & 0.847            & 0.786             & 0.643            \\
 Lite$^3$Pyramid-p3c                                & 0.498           & 0.503            & 0.303           & 0.921          & 0.809           & 0.643          & 0.840            & 0.786             & 0.643            \\
 Meteor                             & 0.744           & 0.601            & 0.424           & 0.901          & 0.762           & 0.571          & 0.796            & 0.714             & 0.500            \\
 MoverScore                         & 0.540           & 0.594            & 0.394           & 0.718          & 0.571           & 0.500          & 0.879            & 0.905             & 0.786            \\
 Novel-1gram            & 0.637           & 0.811            & 0.636           & 0.635          & 0.452           & 0.357          & -0.594           & -0.381            & -0.286           \\
 Novel-2gram            & 0.593           & 0.769            & 0.576           & 0.612          & 0.619           & 0.429          & -0.609           & -0.429            & -0.286           \\
 Repeated-1gram & 0.357           & 0.224            & 0.182           & 0.119          & 0.095           & 0.143          & 0.102            & 0.000             & 0.000            \\
 Repeated-2gram & 0.382           & 0.252            & 0.182           & 0.243          & 0.119           & 0.143          & -0.211           & -0.095            & -0.071           \\
 QAEval-em                          & 0.408           & 0.350            & 0.242           & 0.489          & 0.452           & 0.357          & 0.909            & 0.786             & 0.643            \\
 QAEval-f1                          & 0.602           & 0.489            & 0.333           & 0.588          & 0.500           & 0.429          & 0.894            & 0.714             & 0.571            \\
 ROUGE1                             & \textbf{0.915}           & \textbf{0.881}            & \textbf{0.788}           & 0.704          & 0.476           & 0.357          & 0.909            & 0.786             & 0.643            \\
 ROUGE1p                            & 0.272           & 0.329            & 0.182           & 0.257          & 0.405           & 0.357          & 0.548            & 0.524             & 0.429            \\
 ROUGE1r                            & 0.516           & 0.413            & 0.273           & 0.737          & 0.857           & \textbf{0.714}          & 0.644            & 0.738             & 0.571            \\
 ROUGE2                             & 0.770           & 0.699            & 0.515           & 0.665          & 0.500           & 0.429          & 0.961            & 0.881             & 0.714            \\
 ROUGE2p                            & 0.317           & 0.406            & 0.242           & 0.382          & 0.429           & 0.429          & 0.839            & 0.833             & 0.714            \\
 ROUGE2r                            & 0.651           & 0.510            & 0.364           & 0.893          & 0.786           & 0.643          & 0.842            & 0.786             & 0.643            \\
 ROUGEL                             & 0.811           & 0.790            & 0.606           & 0.620          & 0.429           & 0.429          & 0.897            & 0.857             & 0.714            \\
 ROUGELp                            & 0.275           & 0.280            & 0.151           & 0.243          & 0.357           & 0.286          & 0.660            & 0.667             & 0.571            \\
 ROUGELr                            & 0.600           & 0.524            & 0.364           & 0.815          & 0.809           & 0.643          & 0.688            & 0.738             & 0.571            \\
 SimCSE                             & 0.801           & 0.685            & 0.545           & 0.876          & 0.809           & 0.571          & 0.847            & 0.786             & 0.643            \\
 SummaQA                 & 0.196           & 0.133            & 0.061           & -0.239         & -0.119          & 0.000          & 0.334            & 0.071             & 0.071            \\
 SummaQA-prob                   & 0.591           & 0.503            & 0.212           & 0.251          & 0.214           & 0.143          & 0.416            & 0.357             & 0.286            \\
 Summary-length                     & 0.087           & 0.042            & 0.030           & 0.166          & -0.071          & 0.000          & 0.329            & 0.357             & 0.214            \\
 SUPERT                             & -0.233          & -0.329           & -0.212          & -0.057         & -0.095          & -0.071         & 0.293            & 0.238             & 0.071            \\
 UniEval-coherence                  & -0.620          & -0.357           & -0.273          & 0.062          & 0.095           & 0.071          & 0.481            & 0.333             & 0.286            \\
 UniEval-consistency                & -0.534          & -0.748           & -0.515          & 0.023          & 0.071           & 0.071          & 0.606            & 0.214             & 0.000            \\
 UniEval-fluency                    & 0.286           & 0.189            & 0.121           & -0.681         & -0.643          & -0.500         & 0.667            & 0.500             & 0.357            \\
 UniEval-overall                    & 0.178           & 0.126            & 0.121           & -0.072         & 0.048           & 0.071          & 0.734            & 0.381             & 0.286            \\
 UniEval-relevance                  & 0.365           & 0.762            & 0.545           & 0.064          & 0.333           & 0.214          & 0.717            & 0.548             & 0.429            \\
 \bottomrule
\end{tabular}
\addtolength{\tabcolsep}{1pt} 
\caption{The \textbf{\textit{system-level}} Pearson's $r$, Spearman's $\rho$, and Kendall's $\tau$ correlation coefficients between the automatic metric scores and \textbf{\textit{normalized}} ACU scores of system outputs on CNNDM, XSum and SamSum datasets.
}
\label{tab:metric-corr-sys-normalized-appendix} 
\end{table*}

\begin{table*}[t!]
\small
\centering
\addtolength{\tabcolsep}{-1pt} 
\begin{tabular}{@{\extracolsep{1pt}}lccccccccc@{}}
\toprule
 & \multicolumn{3}{c}{\textbf{CNNDM}} & \multicolumn{3}{c}{\textbf{XSum}} & \multicolumn{3}{c}{\textbf{SamSum}} \\
 % \\
  & $r$ & $\rho$ & $\tau$ &  $r$ & $\rho$ & $\tau$ &  $r$ & $\rho$ & $\tau$ \\
 \cmidrule{2-4} \cmidrule{5-7} \cmidrule{8-10} 
  BARTScore-f1-cnndm                 & 0.398           & 0.384            & 0.296           & 0.276          & 0.274           & 0.228          & 0.445            & 0.413             & 0.341            \\
 BARTScore-f1-parabank              & 0.465           & 0.441            & 0.342           & 0.327          & 0.310           & 0.260          & 0.483            & 0.442             & 0.371            \\
 BARTScore-p-cnndm          & 0.284           & 0.273            & 0.209           & 0.205          & 0.208           & 0.172          & 0.389            & 0.355             & 0.295            \\
 BARTScore-p-parabank       & 0.355           & 0.338            & 0.259           & 0.257          & 0.262           & 0.216          & 0.411            & 0.367             & 0.307            \\
 BARTScore-r-cnndm             & 0.485           & 0.435            & 0.334           & 0.329          & 0.303           & 0.257          & 0.439            & 0.411             & 0.345            \\
 BARTScore-r-parabank          & 0.507           & 0.462            & 0.357           & \textbf{0.361}          & 0.329           & 0.277          & 0.462            & 0.444             & 0.372            \\
 BERTScore-f1-deberta               & 0.518           & 0.491            & 0.386           & 0.317          & 0.323           & 0.274          & \textbf{0.517}            & \textbf{0.472}             & \textbf{0.401}            \\
 BERTScore-f1-roberta               & 0.515           & 0.486            & 0.386           & 0.333          & \textbf{0.330}           & \textbf{0.280}          & 0.512            & 0.470             & 0.401            \\
 BERTScore-p-deberta                & 0.423           & 0.411            & 0.320           & 0.248          & 0.294           & 0.248          & 0.413            & 0.382             & 0.323            \\
 BERTScore-p-roberta                & 0.389           & 0.378            & 0.296           & 0.261          & 0.296           & 0.250          & 0.411            & 0.377             & 0.319            \\
 BERTScore-r-deberta                & 0.491           & 0.454            & 0.354           & 0.329          & 0.291           & 0.246          & 0.474            & 0.441             & 0.372            \\
 BERTScore-r-roberta                & 0.515           & 0.476            & 0.371           & 0.349          & 0.306           & 0.260          & 0.470            & 0.442             & 0.375            \\
 BLANC                              & 0.045           & 0.020            & 0.014           & -0.031         & -0.041          & -0.036         & 0.034            & 0.038             & 0.035            \\
 BLEU                               & 0.441           & 0.414            & 0.328           & 0.294          & 0.300           & 0.260          & 0.469            & 0.432             & 0.368            \\
 CHRF                               & 0.527           & 0.479            & 0.379           & 0.351          & 0.301           & 0.256          & 0.438            & 0.412             & 0.351            \\
 Compression                        & -0.053          & -0.002           & 0.021           & -0.037         & 0.069           & 0.071          & -0.037           & -0.021            & 0.001            \\
 Coverage                           & -0.016          & -0.020           & -0.019          & -0.051         & -0.049          & -0.040         & 0.055            & 0.045             & 0.037            \\
 CTC                                & 0.349           & 0.317            & 0.237           & 0.274          & 0.231           & 0.194          & 0.414            & 0.385             & 0.326            \\
 Density                            & -0.031          & -0.040           & -0.032          & -0.059         & -0.067          & -0.055         & 0.050            & 0.051             & 0.046            \\
 Lite$^3$Pyramid-l2c                                & 0.452           & 0.424            & 0.355           & 0.212          & 0.197           & 0.181          & 0.410            & 0.404             & 0.374            \\
 Lite$^3$Pyramid-l3c                                & 0.449           & 0.427            & 0.358           & 0.213          & 0.197           & 0.180          & 0.418            & 0.417             & 0.385            \\
  Lite$^3$Pyramid-p2c                                & 0.482           & 0.420            & 0.321           & 0.294          & 0.259           & 0.216          & 0.462            & 0.419             & 0.353            \\
 Lite$^3$Pyramid-p3c                                & 0.489           & 0.430            & 0.330           & 0.298          & 0.253           & 0.211          & 0.469            & 0.419             & 0.354            \\
 Meteor                             & 0.484           & 0.435            & 0.337           & 0.329          & 0.303           & 0.260          & 0.427            & 0.391             & 0.335            \\
 MoverScore                         & 0.509           & 0.483            & 0.380           & 0.341          & 0.329           & 0.275          & 0.513            & 0.459             & 0.386            \\
 Novel-1gram            & 0.017           & 0.020            & 0.018           & 0.054          & 0.045           & 0.037          & -0.066           & -0.059            & -0.049           \\
 Novel-2gram            & 0.053           & 0.049            & 0.037           & 0.063          & 0.068           & 0.055          & -0.060           & -0.058            & -0.052           \\
 Repeated-1gram & -0.029          & -0.044           & -0.033          & -0.015         & -0.030          & -0.022         & -0.004           & 0.007             & 0.011            \\
 Repeated-2gram & -0.012          & -0.016           & -0.007          & 0.004          & -0.009          & -0.007         & -0.036           & -0.032            & -0.027           \\
 QAEval-em                          & 0.322           & 0.302            & 0.253           & 0.161          & 0.155           & 0.144          & 0.328            & 0.321             & 0.289            \\
 QAEval-f1                          & 0.412           & 0.379            & 0.301           & 0.227          & 0.207           & 0.188          & 0.367            & 0.349             & 0.307            \\
 ROUGE1                             & \textbf{0.541}           & \textbf{0.510}            & \textbf{0.403}           & 0.324          & 0.324           & 0.278          & 0.494            & 0.464             & 0.399            \\
 ROUGE1p                            & 0.386           & 0.380            & 0.298           & 0.263          & 0.303           & 0.262          & 0.320            & 0.308             & 0.269            \\
 ROUGE1r                            & 0.425           & 0.374            & 0.290           & 0.317          & 0.276           & 0.244          & 0.349            & 0.329             & 0.286            \\
 ROUGE2                             & 0.504           & 0.473            & 0.375           & 0.296          & 0.292           & 0.253          & 0.432            & 0.402             & 0.343            \\
 ROUGE2p                            & 0.439           & 0.424            & 0.335           & 0.270          & 0.291           & 0.253          & 0.346            & 0.330             & 0.283            \\
 ROUGE2r                            & 0.474           & 0.433            & 0.347           & 0.298          & 0.274           & 0.243          & 0.379            & 0.359             & 0.317            \\
 ROUGEL                             & 0.512           & 0.481            & 0.378           & 0.288          & 0.293           & 0.252          & 0.462            & 0.427             & 0.365            \\
 ROUGELp                            & 0.380           & 0.375            & 0.294           & 0.249          & 0.287           & 0.248          & 0.313            & 0.293             & 0.254            \\
 ROUGELr                            & 0.424           & 0.378            & 0.293           & 0.292          & 0.258           & 0.227          & 0.338            & 0.321             & 0.279            \\
 SimCSE                             & 0.437           & 0.393            & 0.300           & 0.321          & 0.305           & 0.255          & 0.454            & 0.422             & 0.356            \\
 SummaQA                 & 0.059           & 0.044            & 0.031           & 0.018          & 0.001           & 0.002          & 0.010            & 0.022             & 0.014            \\
 SummaQA-prob                   & 0.076           & 0.069            & 0.050           & 0.019          & -0.003          & -0.003         & 0.050            & 0.054             & 0.047            \\
 Summary-length                     & 0.040           & 0.002            & -0.021          & 0.027          & -0.075          & -0.079         & 0.010            & 0.021             & -0.001           \\
 SUPERT                             & 0.062           & 0.045            & 0.030           & 0.031          & 0.013           & 0.008          & 0.064            & 0.059             & 0.044            \\
 UniEval-coherence                  & 0.083           & 0.087            & 0.060           & 0.011          & -0.018          & -0.014         & 0.108            & 0.112             & 0.090            \\
 UniEval-consistency                & 0.001           & -0.006           & -0.009          & 0.008          & -0.015          & -0.010         & 0.087            & 0.075             & 0.060            \\
 UniEval-fluency                    & 0.017           & 0.001            & 0.001           & -0.031         & -0.040          & -0.032         & 0.035            & 0.014             & 0.009            \\
 UniEval-overall                    & 0.130           & 0.146            & 0.107           & 0.088          & 0.062           & 0.052          & 0.204            & 0.208             & 0.168            \\
 UniEval-relevance                  & 0.150           & 0.173            & 0.127           & 0.187          & 0.192           & 0.158          & 0.236            & 0.252             & 0.203            \\
 \bottomrule
\end{tabular}
\addtolength{\tabcolsep}{1pt} 
\caption{The \textbf{\textit{summary-level}} Pearson's $r$, Spearman's $\rho$, and Kendall's $\tau$ correlation coefficients between the automatic metric scores and \textbf{\textit{normalized}} ACU scores of system outputs on CNNDM, XSum and SamSum datasets.
}
\label{tab:metric-corr-summ-normalized-appendix} 
\end{table*}

\begin{table*}[t!]
    \small
    \centering
    \addtolength{\tabcolsep}{-2.5pt}
    \begin{tabular}{lrrrrrrrrr}
    \toprule
     & \multicolumn{4}{c}{\textbf{System-level Correlation}} & & \multicolumn{4}{c}{\textbf{Summary-level Correlation}}\\
     % \\
      & \textbf{\preference} & \textbf{\salience} & \textbf{\accuracy} &  \textbf{\textit{n}ACU}& &\textbf{\preference} & \textbf{\salience} & \textbf{\accuracy} &  \textbf{\textit{n}ACU} \\
     \cmidrule{2-5} \cmidrule{7-10} 
    BARTScore\_f1\_cnndm                 & -0.030        & -0.121      & 0.656       & 0.364           &    & 0.060         & 0.032       & 0.335       & 0.280           \\
    BARTScore\_f1\_parabank              & -0.091        & -0.182      & 0.656       & 0.364           &    & 0.079         & 0.038       & 0.369       & 0.324           \\
    BARTScore\_p\_cnndm          & -0.091        & -0.182      & 0.595       & 0.242           &    & -0.008        & -0.033      & 0.281       & 0.210           \\
    BARTScore\_p\_parabank       & -0.273        & -0.364      & 0.534       & 0.242           &    & -0.001        & -0.039      & 0.321       & 0.235           \\
    BARTScore\_r\_cnndm             & 0.545         & 0.455       & -0.076      & 0.212           &    & 0.159         & 0.162       & 0.290       & 0.281           \\
    BARTScore\_r\_parabank          & 0.394         & 0.364       & 0.199       & 0.485           &    & 0.175         & 0.192       & 0.292       & 0.323           \\
    BERTscore\_f1\_deberta               & -0.061        & -0.212      & 0.779       & 0.576           &    & 0.040         & 0.013       & 0.398       & 0.366           \\
    BERTscore\_f1\_roberta               & -0.091        & -0.182      & 0.779       & 0.485           &    & 0.030         & -0.001      & \textbf{0.399}       & 0.358           \\
    BERTscore\_p\_deberta                & -0.333        & -0.485      & 0.626       & 0.364           &    & -0.077        & -0.132      & 0.366       & 0.310           \\
    BERTscore\_p\_roberta                & -0.242        & -0.333      & 0.687       & 0.333           &    & -0.101        & -0.125      & 0.334       & 0.283           \\
    BERTscore\_r\_deberta                & 0.273         & 0.121       & 0.626       & 0.667           &    & 0.171         & 0.178       & 0.312       & 0.314           \\
    BERTscore\_r\_roberta                & 0.273         & 0.121       & 0.565       & \textbf{0.727}           &    & 0.176         & 0.193       & 0.331       & 0.328           \\
    BLANC                              & 0.545         & 0.576       & -0.504      & -0.212          &    & 0.256         & 0.310       & -0.051      & -0.006          \\
    BLEU                               & -0.182        & -0.333      & 0.534       & 0.515           &    & -0.025        & -0.046      & 0.260       & 0.296           \\
    CHRF                               & 0.576         & 0.424       & 0.199       & 0.485           &    & 0.161         & 0.181       & 0.284       & 0.347           \\
    Compression                        & -0.606        & -0.758      & 0.382       & 0.091           &    & -0.344        & -0.422      & 0.081       & 0.046           \\
    Coverage                           & -0.030        & 0.121       & -0.779      & -0.606          &    & 0.076         & 0.074       & -0.035      & -0.019          \\
    CTC                                & 0.455         & 0.545       & -0.473      & -0.242          &    & 0.267         & 0.295       & 0.193       & 0.221           \\
    Density                            & 0.121         & 0.273       & -0.687      & -0.515          &    & 0.134         & 0.117       & -0.082      & -0.055          \\
    Lite$^3$Pyramid-l2c                                & 0.545         & 0.576       & -0.046      & 0.212           &    & 0.214         & 0.218       & 0.230       & 0.337           \\
    Lite$^3$Pyramid-l3c                                & 0.545         & 0.636       & -0.107      & 0.091           &    & 0.199         & 0.219       & 0.223       & 0.342           \\
    Lite$^3$Pyramid-p2c                                & 0.576         & 0.667       & -0.168      & 0.121           &    & 0.234         & 0.254       & 0.206       & 0.301           \\
    Lite$^3$Pyramid-p3c                                & 0.576         & 0.667       & -0.107      & 0.182           &    & 0.210         & 0.242       & 0.196       & 0.305           \\
    Meteor                             & 0.394         & 0.242       & 0.382       & 0.485           &    & 0.157         & 0.149       & 0.274       & 0.313           \\
    MoverScore                         & -0.151        & -0.364      &\textbf{ 0.870}       & 0.545           &    & 0.009         & -0.018      & 0.360       & 0.356           \\
    Novel-1gram             & 0.030         & -0.121      & 0.779       & 0.606           &    & -0.071        & -0.066      & 0.037       & 0.018           \\
    Novel-2gram            & -0.061        & -0.273      & \textbf{0.870}       & 0.636           &    & -0.097        & -0.088      & 0.088       & 0.057           \\
    Repeated-1gram & 0.212         & 0.303       & -0.321      & -0.121          &    & 0.097         & 0.102       & -0.035      & -0.032          \\
    Repeated-2gram & 0.000         & -0.091      & -0.046      & 0.030           &    & 0.068         & 0.074       & -0.010      & -0.007          \\
    QAEval-em                          & 0.303         & 0.333       & -0.107      & 0.030           &    & 0.087         & 0.100       & 0.131       & 0.227           \\
    QAEval-f1                          & 0.485         & 0.515       & -0.076      & 0.151           &    & 0.127         & 0.122       & 0.203       & 0.274           \\
    ROUGE1                             & -0.061        & -0.212      & 0.840       & 0.636           &    & 0.033         & -0.008      & 0.346       & \textbf{0.377}           \\
    ROUGE1p                            & -0.364        & -0.515      & 0.687       & 0.394           &    & -0.158        & -0.241      & 0.293       & 0.278           \\
    ROUGE1r                            & \textbf{0.697}         & 0.667       & -0.107      & 0.182           &    & 0.272         & 0.322       & 0.211       & 0.248           \\
    ROUGE2                             & 0.000         & -0.151      & 0.595       & 0.636           &    & 0.004         & -0.005      & 0.304       & 0.329           \\
    ROUGE2p                            & -0.273        & -0.424      & 0.626       & 0.424           &    & -0.095        & -0.118      & 0.309       & 0.302           \\
    ROUGE2r                            & 0.455         & 0.364       & 0.199       & 0.424           &    & 0.134         & 0.177       & 0.248       & 0.285           \\
    ROUGEL                             & -0.061        & -0.212      & 0.779       & 0.636           &    & 0.024         & 0.005       & 0.325       & 0.370           \\
    ROUGELp                            & -0.394        & -0.545      & 0.656       & 0.364           &    & -0.148        & -0.218      & 0.279       & 0.289           \\
    ROUGELr                            & 0.606         & 0.576       & 0.046       & 0.333           &    & 0.234         & 0.296       & 0.211       & 0.263           \\
    SimCSE                             & 0.273         & 0.242       & 0.351       & 0.364           &    & 0.154         & 0.154       & 0.266       & 0.284           \\
    SummaQA                & 0.636         & 0.606       & -0.290      & 0.000           &    & 0.158         & 0.218       & -0.020      & 0.000           \\
    SummaQA-prob                   & 0.515         & 0.424       & 0.260       & 0.303           &    & 0.144         & 0.176       & 0.021       & 0.049           \\
    Summary-length                     & 0.576         & \textbf{0.727}       & -0.473      & -0.182          &    & \textbf{0.344}         & \textbf{0.422}       & -0.081      & -0.046          \\
    SUPERT                             & 0.333         & 0.485       & -0.656      & -0.424          &    & 0.206         & 0.245       & -0.026      & 0.029           \\
    UniEval-coherence                  & 0.151         & 0.121       & -0.107      & -0.061          &    & 0.229         & 0.176       & 0.025       & 0.071           \\
    UniEval-consistency                & -0.091        & 0.061       & -0.565      & -0.545          &    & 0.077         & 0.064       & -0.045      & -0.030          \\
    UniEval-fluency                    & 0.303         & 0.273       & 0.260       & 0.333           &    & 0.080         & 0.058       & 0.004       & 0.018           \\
    UniEval-overall                    & 0.151         & 0.000       & 0.443       & 0.303           &    & 0.219         & 0.142       & 0.116       & 0.122           \\
    UniEval-relevance                  & -0.061        & -0.212      & 0.656       & 0.394           &    & 0.191         & 0.114       & 0.181       & 0.129           \\
    \bottomrule
    \end{tabular}
     \addtolength{\tabcolsep}{2.5pt}
    \caption{The Kendall's correlation between the automatic metric and different human evaluation protocols on CNNDM dataset.
    }
    \label{tab:metric-corr-protocol-appendix} 
    \end{table*}
\end{document}